\def\year{2022}\relax
\documentclass[letterpaper]{article} 
\usepackage{aaai22}  
\usepackage{times}  
\usepackage{helvet}  
\usepackage{courier}  
\usepackage[hyphens]{url}  
\usepackage{graphicx} 
\usepackage{natbib}  
\usepackage{caption} 
\DeclareCaptionStyle{ruled}{labelfont=normalfont,labelsep=colon,strut=off} 
\usepackage{algorithm}

\usepackage{newfloat}
\usepackage{listings}

\usepackage{epsfig}
\usepackage{amsmath}
\usepackage{amssymb}

\usepackage{xspace}
\usepackage{algpseudocode}
\usepackage{bm}
\usepackage{color}
\usepackage{comment}
\usepackage[normalem]{ulem}
\usepackage{makecell}
\usepackage{multirow}
\usepackage{subfigure}
\usepackage{booktabs}
\usepackage{xcolor,colortbl}
\usepackage{float}
\usepackage{mathtools}
\definecolor{Gray}{gray}{0.9}
\definecolor{LightCyan}{rgb}{0.88,1,1}

\newtheorem{remark}{Remark}

\def\onedot{\ifx\@let@token.\else.\null\fi\xspace}
\def\eg{\emph{e.g}\onedot} 
\def\eg{\emph{e.g}\onedot} 
\def\ie{\emph{i.e}\onedot}

\def\wrt{w.r.t\onedot} 
\def\etal{\emph{et al}\onedot}
\makeatother

\def\Vec#1{{\boldsymbol{#1}}}
\def\Mat#1{{\boldsymbol{#1}}}

\DeclareMathOperator*{\argmin}{arg\,min}
\newcommand\correspondingauthor{\thanks{Corresponding author.}}

\floatstyle{ruled}
\newfloat{listing}{tb}{lst}{}
\floatname{listing}{Listing}
\title{Adaptive Poincar{\'e} Point to Set Distance for Few-Shot Classification}
\author{
    Rongkai Ma\textsuperscript{\rm 1},
    Pengfei Fang\textsuperscript{\rm 2,}\textsuperscript{\rm 3}\correspondingauthor,
    Tom Drummond\textsuperscript{\rm 4},
    Mehrtash Harandi\textsuperscript{\rm 1,}\textsuperscript{\rm 3}
}
\affiliations{
    \textsuperscript{\rm 1}Monash University,
    \textsuperscript{\rm 2}The Australian National University,
    \textsuperscript{\rm 3}DATA61-CSIRO, Australia,
    \textsuperscript{\rm 4}The University of Melbourne\\
    \tt\small \{rongkai.ma, mehrtash.harandi\}@monash.edu,
    \tt\small pengfei.fang@anu.edu.au,
    \tt\small tom.drummond@unimelb.edu.au 
    
}

\begin{document}
\maketitle

\begin{abstract}
Learning and generalizing from limited examples, \ie, few-shot learning, is of core importance to many real-world vision applications. A principal way of achieving few-shot learning is to realize an embedding where samples from different classes are distinctive. Recent studies suggest that embedding via hyperbolic geometry enjoys low distortion for hierarchical and structured data, making it suitable for few-shot learning. In this paper, we propose to learn a context-aware hyperbolic metric to characterize the distance between a point and a set associated with a learned set to set distance. To this end, we formulate the metric as a weighted sum on the tangent bundle of the hyperbolic space and develop a mechanism to obtain the weights adaptively and based on the constellation of the points. This not only makes the metric local but also dependent on the task in hand, meaning that the metric  will adapt depending on the samples that it compares. We empirically show that such metric yields robustness in the presence of outliers and achieves a tangible improvement over baseline models. This includes the state-of-the-art results on five popular few-shot classification benchmarks, namely \emph{mini}-ImageNet, \emph{tiered}-ImageNet, Caltech-UCSD Birds-200-2011 (CUB), CIFAR-FS, and FC100.

\end{abstract}

\section{Introduction}
In the modern context of machine learning, deep neural networks (DNNs) have enjoyed enormous success by leveraging the rich availability of labeled data for supervised training. Despite this, deep supervised learning is primarily limited in terms of scaling towards unseen samples due to the high cost of acquiring large amounts of labeled data. This is in clear contrast to how humans learn, where in many cases, only a handful of training examples are sufficient for generalizing towards unseen samples. Few-Shot Learning (FSL) addresses this critical problem through the development of algorithms that can learn using limited data~\cite{finn2017model, nichol2018first, snell2017prototypical,sung2018learning,vinyals2016matching,ye2020few,Hong_2021_CVPR_RAP,wang2020instance}.

\begin{figure}[t]
\centering
	\subfigure[]{\includegraphics[width=0.45\linewidth]{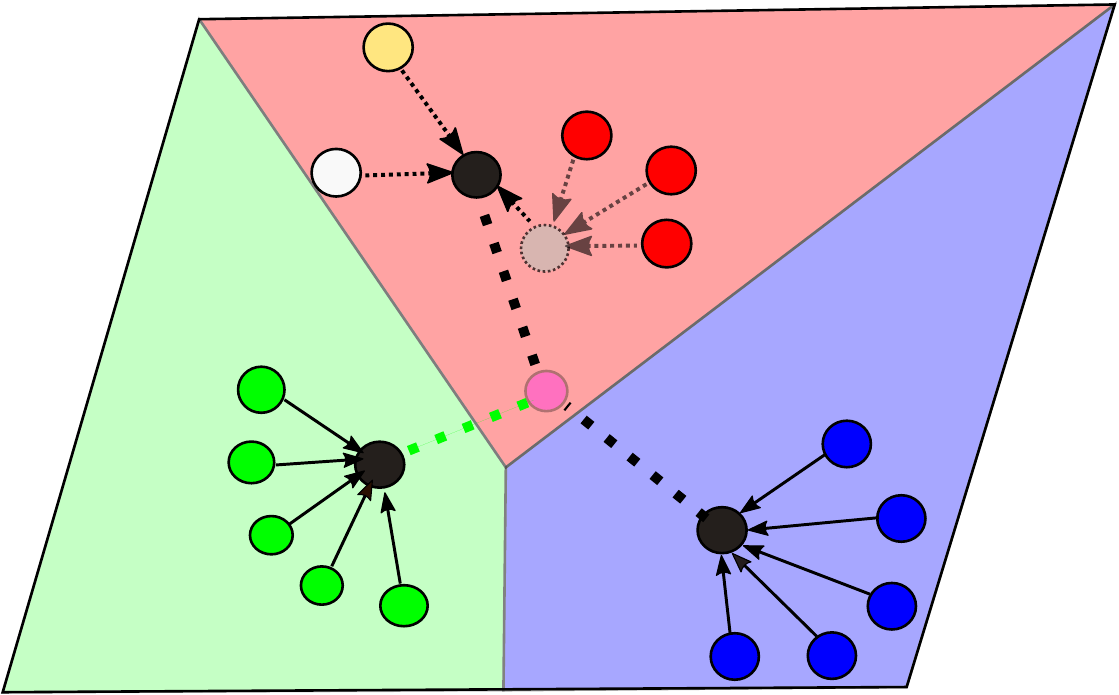}\label{fig:protonet intro}}%
	\hfil
	\subfigure[]{\includegraphics[width=0.45\linewidth]{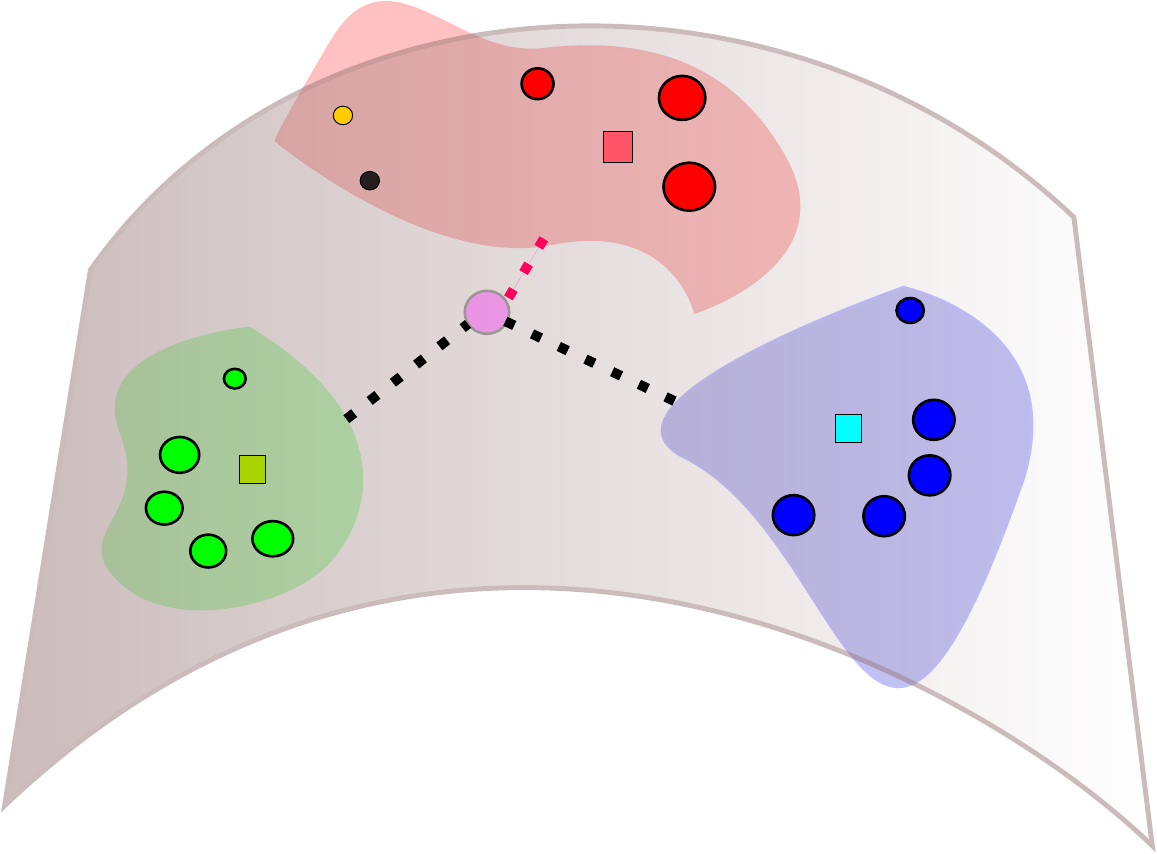}\label{fig:p2s intro}}%
	\hfil
	\caption{(a): For a query sample in ``red" class, outliers (\ie, yellow  and white circles) drag the  prototype (\ie, the black circle) far away from the real cluster center in the embedding space such that the nearest neighbor classifier mis-classifies the query point into ``green" class. (b): Our method computes an adaptive point to set distance on the manifold, which is more robust to outliers than prototypes. Best viewed in color.
	}\label{fig:compare}
\end{figure}

Performing FSL well is essential towards creating robust frameworks that can learn with the efficiency of humans. In many cases,  FSL methods deem to learn an embedding space to distinguish samples from different classes. Therein, the embedding space is a multidimensional Euclidean space and is realized via a deep neural network. 

Employing hyperbolic geometry to encode data has been shown rewarding, as the volume of  space expands exponentially~\cite{ganea2018hyperbolic, khrulkov2020hyperbolic}. Recent works have shown that a hierarchical structure exists within visual datasets and that the use of hyperbolic embeddings can yield significant improvements over Euclidean embeddings~\cite{khrulkov2020hyperbolic,Fang_2021_ICCV}.

Most existing FSL solutions learn a metric through comparing the distance between a query sample and the class prototypes, often modeled as the mean embeddings of each class. However, this does not take the adverse effects of outliers and noises into consideration~\cite{sun2019hierarchical}. This severely limits the representation power of embedding-based methods since the outliers may drag the prototype away from the true center of the cluster (see Fig.\ref{fig:protonet intro}). For a more robust approach, we require an adaptive metric, which can faithfully capture the distribution per class, while being robust to outliers and other nuances in data (Fig.~\ref{fig:p2s intro}). 

With this in mind, we propose learning a context-aware hyperbolic metric that characterizes the point to set (dis)similarities. This is achieved through employing a Poincar\'e ball to model hyperbolic spaces and casting the (dis)similarity as a weighted-sum between a query and a class that is learned adaptively.
In doing so, each sample (from the support and query sets) is modeled by a set itself (\ie, a feature map). Therefore, we propose to make use of pairwise distances between elements of two sets, along with a refinement mechanism to disregard uninformative parts of the feature maps. 
This leads to a flexible and robust framework for the FSL tasks. We summarize our contributions as follows: 

\begin{itemize}
    \item We propose a novel adaptive Poincar\'e point to set (APP2S) distance metric for the FSL task. 
    \item We further design a mechanism to produce a weight, dependent on the constellation of the point, for our APP2S metric.
    \item We conduct extensive experiments across five FSL benchmarks to evaluate the effectiveness of the proposed method.
    \item We further study the robustness of our method, which shows our method is robust against the outliers compared to competing baselines.
\end{itemize}

\section{Preliminaries}
\label{preliminary}

In what follows, we use $\mathbb{R}^n$ and $\mathbb{R}^{m \times n}$ to denote the $n$-dimensional Euclidean space and space of $m \times n$ real matrices, respectively. The $n$-dimensional hyperbolic space is denoted by $\mathbb{H}_c^n$. The $\mathrm{arctanh}: (-1,1) \to \mathbb{R}, \mathrm{arctanh}(x) = \frac{1}{2}\ln(\frac{1 + x}{1 - x}), |x|<1$ refers to the inverse hyperbolic tangent function. 
The vectors and matrices (or 3-D tensors) are denoted by bold lower-case letters and bold upper-case letters throughout the paper.

\subsection{Riemannian Geometry}
In this section, we will give a brief recap of Riemannian geometry.
A manifold, denoted by $\mathcal{M}$, is a  curved surface, which locally resembles the Euclidean space.
The tangent space at $\Vec{x} \in \mathcal{M}$ is denoted by $T_{\Vec{x}} \mathcal{M}$. It contains all possible vectors passing through point $\Vec{x}$ tangentially. On the manifold, the shortest path connecting two points is a geodesic, and its length is used to measure the distances on the manifold.

\subsection{Hyperbolic Space}
Hyperbolic spaces are Riemannian manifolds with constant negative curvature and can be studied using the 
Poincar\'e ball model~\cite{ganea2018hyperbolic,khrulkov2020hyperbolic}.
The Poincar\'e ball ($\mathbb{D}_c^n, g^{c}$) is  a smooth $n$-dimensional manifold identified by satisfying $\mathbb{D}^n_c = \{\Vec{x} \in \mathbb{R}^n: c\lVert \Vec{x} \rVert < 1 , c\geqslant0\}$\footnote{In the supplementary material, we  provide further details regarding  the Poincar{\'e} ball model and its properties.}, where $c$ is the absolute value of the curvature for a Poincar{\'e} ball, while the real curvature value is $-c$. The Riemannian metric $g^{c}$ at $\Vec{x}$ is defined as $g^{c} = \lambda{_{\Vec{x}}^c}^2g^{E}$, where $g^E$ is the Euclidean metric tensor and $\lambda_{\Vec{x}}^c$ is the conformal factor, defined as:
\begin{equation}
\lambda_{\Vec{x}}^c\coloneqq\frac{2}{1- c\lVert \Vec{x} \rVert^2}.     
\end{equation}

Since the hyperbolic space is a non-Euclidean space, the rudimentary operations, such as vector addition, cannot be applied (as they are not faithful to the geometry). The M{\"o}bius gyrovector space provides many standard operations for hyperbolic spaces. Essential to our developments in this work is the M{\"o}bius addition of two points $\Vec{x}, \Vec{y} \in \mathbb{D}^n_c$, which is calculated as:
\begin{equation}\label{eq:mobius addition}
\Vec{x} \oplus_{c} \Vec{y} = \frac{(1+2c \langle \Vec{x},\Vec{y} \rangle + c\|\Vec{y} \|^2)\Vec{x}+(1-c\|\Vec{x}\|^2 )\Vec{y}}{1+2c\langle \Vec{x},\Vec{y} \rangle+c^2\|\Vec{x}\|^2 \|\Vec{y}\|^2}.
\end{equation}
The geodesic distance between two points $\Vec{x}, \Vec{y} \in \mathbb{D}^n_c$ can be obtained as:
\begin{equation}\label{eq:distance}
d_c(\Vec{x},\Vec{y})=\frac{2}{\sqrt{c}}\text{arctanh}(\sqrt{c}\|-\Vec{x}\oplus_c \Vec{y}\|).
\end{equation} 

Another essential operation used in our model is the hyperbolic averaging. The counterpart of Euclidean averaging in hyperbolic space is the $E\emph{instein mid-point}$ which has the most simple form in $K\emph{lein}$ coordinates (another model of the hyperbolic space which is isometric to the Poincar{\'e} ball). 
Thus, we transform the points from Poincar{\'e} (\ie, $\Vec{x}_{\mathbb{D}}$) ball model to Klein model (\ie, 
$\Vec{x}_{\mathbb{K}}$) using the transformation:
\begin{equation}\label{eq: p2k}
    \Vec{x}_{\mathbb{K}} = \frac{2\Vec{x}_{\mathbb{D}}}{1+c\|\Vec{x}_{\mathbb{D}}\|^2}.
\end{equation}
Then, the hyperbolic averaging in Klein model is obtained as:
\begin{equation}\label{eq: hype ave}
    \text{HypAve}(\Vec{x}_1,\ldots,\Vec{x}_{N})=\sum_{i=1}^N\gamma_i\Vec{x}_i/\sum_{i=1}^N\gamma_i,
\end{equation}
where $\gamma_i=\frac{1}{\sqrt{1-c\|\Vec{x}_i\|^2}}$ are the Lorentz factors. Finally, we transform the coordinates back to Poincar{\'e} model using:
\begin{equation}\label{eq: k2p}
    \Vec{x}_{\mathbb{D}} = \frac{\Vec{x}_{\mathbb{K}}}{1+\sqrt{1-c\|\Vec{x}_{\mathbb{K}}\|^2}}.
\end{equation}

In our work, we make use of the tangent bundle  of the $\mathbb{D}^n_c$. 
The logarithm map defines a function from $\mathbb{D}^n_c \to T_{\Vec{x}}\mathbb{D}_c^n$, which projects a point in the Poincar\'e ball onto the tangent space at $\Vec{x}$, as:
\begin{equation}
    \label{eq:logmap}
    \Vec{\pi}^c_\Vec{x}(\Vec{y})=\frac{2}{\sqrt{c}\lambda^c_\Vec{x}}\text{arctanh}(\sqrt{c}\|-\Vec{x}\oplus_c\Vec{y}\|)\frac{-\Vec{x}\oplus_c\Vec{y}}{\|-\Vec{x}\oplus_c\Vec{y}\|}.
\end{equation}

\subsection{Point to Set Distance}
Let $\mathcal{S} = \{\Vec{s}_1, \ldots, \Vec{s}_k\}$ be a set. The distance from a point $\Vec{p}$ to the set $\mathcal{S}$ can be defined in various forms. The min and max distance from a point $\Vec{p}$ to the set $\mathcal{S}$ are two simple metrics, which can be defined as: 
\begin{equation}\label{eq:leastdistance}
d_{\mathrm{p2s}}^{\mathrm{l}} (\Vec{p}; \mathcal{S})=\inf \{d(\Vec{p}, \Vec{s}_i) | \Vec{s}_i \in \mathcal{S}\},     
\end{equation}
\begin{equation}\label{eq:highestdistance}
d_{\mathrm{p2s}}^{\mathrm{h}} (\Vec{p}; \mathcal{S})=\sup \{d(\Vec{p}, \Vec{s}_i) | \Vec{s}_i \in \mathcal{S}\}, 
\end{equation}
where $\inf$ and $\sup$ are the infimum and supremum functions, respectively. Given their geometrical interpretation, $d_{\mathrm{p2s}}^{\mathrm{l}}$ and $d_{\mathrm{p2s}}^{\mathrm{h}}$ 
define the lower and upper pairwise bounds, and fail to encode structured  information about the set. Therefore, we opt for a weighted-sum formalism to measure the distance between a point and a set in~\S~\ref{sec:adaptive dist}.

\section{Method}
This section will give an overview of the proposed method, followed by a detailed description of each component in our model.

\begin{figure*}[t]
\centering
	\subfigure[]{\includegraphics[width=0.79\linewidth, height = 6cm]{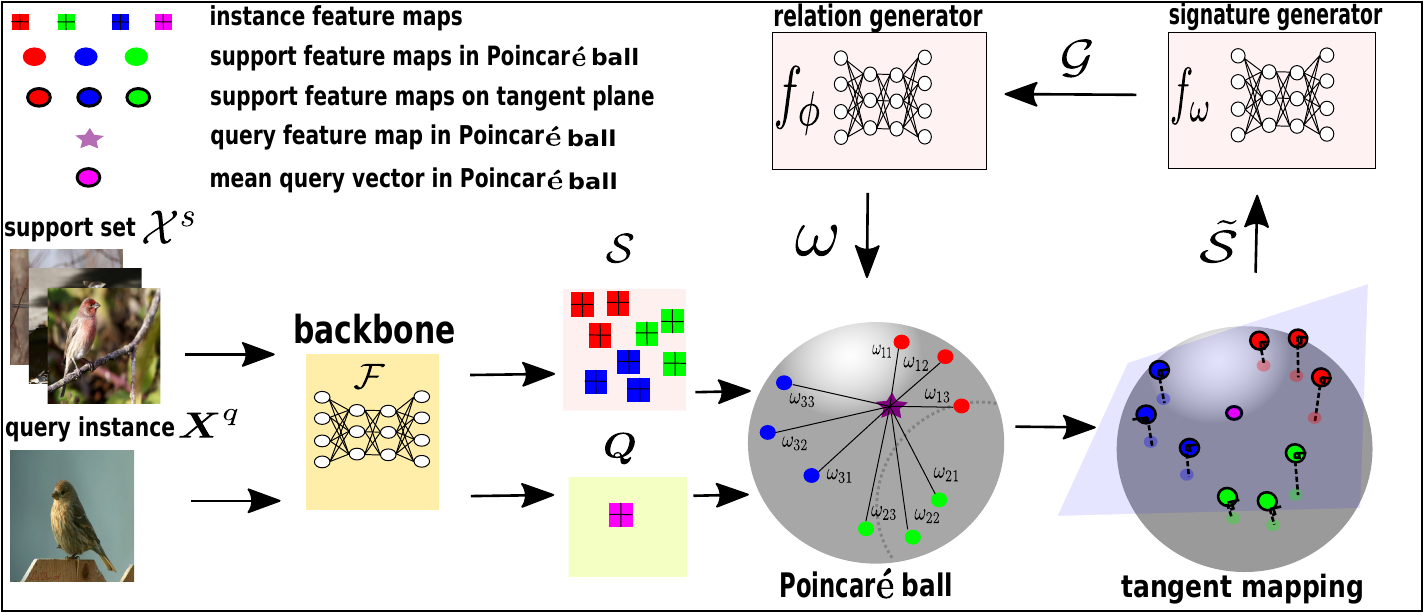}\label{fig:framework}}%
	\hfil
	\subfigure[]{\includegraphics[width=0.18\linewidth, height = 6cm]{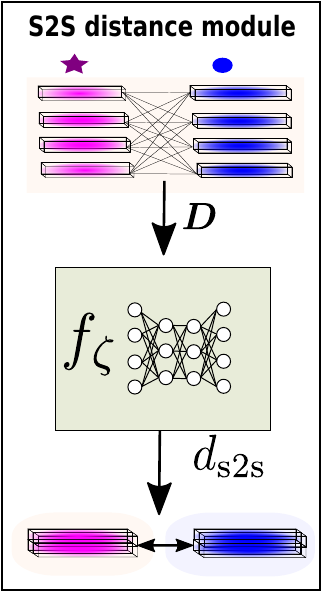}\label{fig:s2s}}%
	\hfil
	\caption{(a): The overall pipeline of our method. Given an episode,  we use a backbone network to extract and map the inputs to a hyperbolic space. We then project the support samples onto the tangent plane of the query point and employ a  refinement function $f_{\omega}$ to obtain the class and episode-aware signature of every class in the support set. This is followed by a mapping $f_{\phi}$ that weighs the importance of each sample of the support set \wrt their corresponding class. This enables us to calculate an adaptive point to set distance towards inference. (b): The S2S distance module. We calculate a pair-wise distance between each feature vector of the two feature maps as the input of the distance network $f_{\zeta}$. Then the $f_{\zeta}$ outputs a S2S distance, which is further used to compute the adaptive P2S.
	}\label{fig:pipeline}
\end{figure*}

\subsection{Problem Formulation}
We follow the standard protocol to formulate few-shot learning (FSL) with episodic training. An episode represents an $N$-way $K$-shot classification problem (\ie, the training set, named support set, includes $N$ classes where each class has $K$ examples). 
As the name implies, $K$ (\ie, the number of examples per class) is small (\eg, $K = 1$ or $5$). 
The goal of learning is to realize a function $\mathcal{F}: \mathcal{X} \to \mathbb{R}^n$ to embed the support set
to a latent and possibly lower-dimensional space, such that query samples can be recognized easily using a nearest neighbour classifier. To be specific, an episode or task $\mathcal{E}_i$ consists of a query set  $\mathcal{X}^q = \{(\Vec{X}^q_{i}, y^q_i) |  i = 1, \ldots, N\}$, where $\Vec{X}^q_{i}$ denotes a query example\footnote{Without losing generality, we use one sample per class as a query for presenting our method. In practice, each episode contains multiple samples for the query image per class.} sampled from class $y^q_i$, and a support set  $\mathcal{X}^s = \{ (\Vec{X}^s_{ij}, y^s_{i})| i = 1, \ldots, N, j = 1, \ldots, K\}$, where $\Vec{X}^s_{ij}$ denotes the $j$-th sample in the class $y^s_i$. 
The embedding methods for FSL, our solution being one, often formulate training as:
\begin{equation}\label{eq:obj}
\mathcal{F}^* : = \argmin_{\mathcal{F}}    \sum_{\Mat{X}^q_{u}\in\mathcal{X}^q}\delta\big( \mathcal{F}(\Vec{X}^q_u), \mathcal{F}({\mathcal{X}}^s_v) \big)~~\text{s.t.}~y^q_u = y^s_v, 
\end{equation}
where $\delta$ measures a form of distance between the query and the support samples.



\subsection{Model Overview}
We begin by providing a sketch of our method (see the conceptual diagram in Fig.~\ref{fig:framework} and Fig.~\ref{fig:s2s}). The feature extractor network, denoted by $\mathcal{F}$, maps the input to a hyperbolic space in our work. We model every class in the support set by its signature. The signature is both class and episodic-aware, meaning that the signature will vary if the samples of a class or samples in the episode vary. This will enable us to calculate an adaptive distance from the query point to every support-class while being vigilant to the arrangement and constellation of the support samples. 
We stress that our design is different from many prior works where class-specific prototypes are learned for FSL.
For example, in~\cite{ khrulkov2020hyperbolic,snell2017prototypical,sung2018learning}, the prototypes are class-specific but not necessarily episodic-aware.

To obtain the signatures for each class in the support set, we project the support samples onto the tangent space of the query point and feed the resulting vectors to a signature generator $f_{\omega}$. 
The signature generator realizes a permutation-invariant function and  refines and summarizes its inputs  to one signature per class. We then leverage a relational network $f_{\phi}$ to contrast samples of a class against their associated signature and produce a relational score. To obtain  the adaptive P2S distance, we first compute a set to set (S2S) distance between the query feature map and each support feature map using the distance module $f_{\zeta}$. Moreover, a weighted-sum is calculated using the relational score acting as the weight on the corresponding S2S distance, which serves as the P2S distance.  


Given P2S distances, our network is optimized by minimizing the adaptive P2S distance between the query and its corresponding set while ensuring that the P2S distance to other classes (\ie, wrong classes) is maximized.

\subsection{Adaptive Poincar\'e Point to Set Distance}\label{sec:adaptive dist}

In FSL, we are given a small support set of $K$  images, $\mathcal{X}^s_i = \{\Vec{X}^s_{i1}, \ldots, \Vec{X}^s_{iK}\}$ per class $y^s_{i}$ to learn a classification model. We use a deep neural network to first encode the input to a multi-channel feature map, as $\mathcal{S}_i = \mathcal{F}(\mathcal{
X}_i^s)$, with $\mathcal{S}_i = \{\Mat{S}_{i1}, \ldots, \Mat{S}_{iK} |\Mat{S}_{ij} \in \mathbb{R}^{H\times W\times C} \}$, where $H$, $W$, and $C$ indicate the height, width, and channel size of the instance feature map. Each feature map consists of a set of patch descriptors (local features), which can be further represented as $\Mat{S}_{ij}=\{\Vec{s}_{ij}^{1},\ldots,\Vec{s}_{ij}^{HW} | \Vec{s}_{ij}^r \in \mathbb{R}^C \}$.

In our work, we train the network to embed the representation in the Poincar\'e ball; thus, we need to impose a constraint on patch descriptors at each spatial location $\Vec{s}_{ij}^{r}$ as follows:
\begin{equation}
\label{eq: clip norm implement}
\Vec{s}_{ij}^{r}=\left\{\begin{matrix}
\Vec{s}_{ij}^{r} &~\text{if}~\|\Vec{s}_{ij}^{r}\| \leq \mu
\\ 
\mu \Vec{s}_{ij}^{r} / \|\Vec{s}_{ij}^{r}\| &~~\text{if}~\|\Vec{s}_{ij}^{r}\| > \mu,
\end{matrix}\right.
\end{equation}
where $\mu$ is the norm upper bound of the vectors in the Poincar{\'e} ball. In our model, we choose $\mu=(1-\epsilon)/c$, where $c$ is the curvature of the Poincar\'e ball and $\epsilon$ is a small value 
that makes the system numerically stable. The same operation applies to the query sample, thereby obtaining an instance feature map for the query sample $\Mat{Q} = \{\Vec{q}^{1}, \ldots, \Vec{q}^{HW}\}, \Vec{q}^{r}\in\mathbb{D}^C_c$. Then the P2S distance between the query sample $\Mat{Q}$ and the support set per class $\mathcal{S}_i$ can be calculated using Eq.~\eqref{eq:leastdistance} or Eq.~\eqref{eq:highestdistance}. However, those two metrics only determine the lower or upper bound of P2S distance, thereby ignoring the structure and  distribution of the set to a great degree. To make better use of the distribution of samples in a set, we propose the adaptive P2S distance metric as: 
\begin{equation}\label{eq:adaptive}
d_{\mathrm{p2s}}^{\mathrm{adp}}(\Mat{Q}; {\mathcal{S}}_i) \coloneqq \frac{\sum\limits_{j = 1}^{K} w_{ij}d(\Mat{Q}, {\Mat{S}}_{ij})}
{\sum\limits_{j = 1}^K w_{ij}},   
\end{equation}
where $w_{ij}$ is the adaptive factor for $d(\Mat{Q}, {\Mat{S}}_{ij})$. We refer to the distance in Eq.~\eqref{eq:adaptive}  as Adaptive Poincar\'e Point to Set (APP2S) distance, hereafter.

In Eq.~\eqref{eq:adaptive}, we need to calculate the distance between two feature maps  (\ie, $d(\Mat{Q}, \Mat{S}_{ij})$). In doing so, we formulate a feature map as a set (\ie, $\{\Vec{q}^{1},\ldots,\Vec{q}^{HW}\}$ and $\{\Vec{s}_{ij}^{1},\ldots,\Vec{s}_{ij}^{HW}\}$), such that a set to set (S2S) distance can be obtained. One-sided Hausdorff and two-sided Hausdorff distances~\cite{huttenlocher1993comparing} are two widely used metrics to measure the distance between sets. However, these two metrics are sensitive to outliers~\cite{huttenlocher1993comparing,ibanez2008use}. To alleviate this issue, we propose to learn the S2S distance by a network  $f_{\zeta}$. We first calculate the pair-wise distance between two sets as $\Mat{D}(\Mat{Q}, \Mat{S}_{ij}) \in \mathbb{R}^{HW \times HW}$, where each element in $\Mat{D}$ is obtained by $d_{h, w} = d_c(\Vec{q}^h, \Vec{s}_{ij}^w )$ using Eq.~\eqref{eq:distance}, where $h = 1, \ldots, HW$ and $w = 1, \ldots, HW$. Then we use a neural network to further learn the distance between two feature maps (see Fig.~\ref{fig:s2s}), which is given by:
\begin{equation}\label{eq: s2s}
d_{\mathrm{s2s}}(\Mat{Q}, \Mat{S}_{ij}) : = f_{\zeta}(\Mat{D}). 
\end{equation}
Comparing to the Hausdorff distance~\cite{Conci2018DistanceBS} (see supplementary material), our set to set distance is more flexible and learned through the optimization process.

To further obtain the weights of APP2S (\ie, $w_{ij}$), we make use of the tangent space of the query sample. We first compute a mean query vector $\Vec{\bar{q}}$ over the spatial dimensions of the feature map $\Mat{Q}$ using Eq.~\eqref{eq: p2k}- Eq.~\eqref{eq: k2p}. Then, we project the samples in the support set to the tangent space of the mean query vector (see Fig.~\ref{fig:framework}), using the logarithm map as: 
\begin{equation}
\tilde{\mathcal{S}}  = \Vec{\pi}^c_{\bar{\Vec{q}}}({\mathcal{S}}),
\end{equation}
where $\tilde{\mathcal{S}}$ indicates the projected support set on the tangent space at $\bar{\Vec{q}}$. For the $i$-th class, we can obtain a set of feature maps: $\tilde{\mathcal{S}}_i = \{\tilde{\Mat{S}}_{i1}, \ldots, \tilde{\Mat{S}}_{iK} \}$\footnote{The projected feature map is also composed by the vectors at each spatial location $\tilde{\Mat{S}}_{ij}=\{\tilde{\Vec{s}}_{ij}^{{1}},\ldots, \tilde{\Vec{s}}_{ij}^{{HW}}\}$}. To obtain a meaningful weight $w_{ij}$, we first propose a signature generator, which jointly refines sample representations in the support set and summarizes the set representation per class as the class signature. As shown in Fig.~\ref{fig:framework}, the signature generator receives the projected support set $\tilde{\mathcal{S}}= \{\tilde{\mathcal{S}}_{1}, \ldots, \tilde{\mathcal{S}}_{N} \}$ as input and refines them for the follow-up task (\ie, obtaining the weights for the APP2S). We denote the output of the refinement module by  $ \hat{\mathcal{S}}= f_{\omega}(\tilde{\mathcal{S}})$ ($\hat{\mathcal{S}} = \{\hat{\mathcal{S}}_{1}, \ldots,  \hat{\mathcal{S}}_{N} \}$, $\hat{\mathcal{S}}_{i} = \{\hat{\Mat{S}}_{i1}, \ldots, \hat{\Mat{S}}_{iK} \}$). One can understand the refinement function as learning  the context of the support set by seeing all the samples, thereby highlighting the discriminative samples and restraining the non-informative samples such as outliers for all the samples. Then the signature for each class is obtained by summarizing as: $\bar{\Vec{S}}_{i} ={ \sum_{j=1}^{K}\hat{\Mat{S}}_{ij}}/{K}$.
\begin{remark}
Our proposed set-signature generator $f_{\omega}$ is similar to the set-to-set function in FEAT~\cite{ye2020few}, in the sense that both functions perform self-attention over the input features. However, the fundamental difference is that our module exploits the relation between the spatial feature descriptors of all samples in a support set~(\eg, $\tilde{\Vec{s}}^{r}_{ij}$), instead of prototypes as proposed in FEAT~\cite{ye2020few}, which possibly gives the model more flexibility to encode meaningful features.
\end{remark}

Given sample features in a class $\tilde{\mathcal{S}}_i = \{\tilde{\Mat{S}}_{i1}, \ldots, \tilde{\Mat{S}}_{iK} \}$ and the corresponding class signature $\bar{\Mat{S}}_{i}$, we use a relation generator (\ie, $f_{\phi}$ in Fig.~\ref{fig:framework}) to compare the relationship between an individual feature map and the class signature. In doing so, we first concatenate the individual feature maps and their class signature along the channel dimension to obtain a hybrid representation, as:
\begin{equation}
\Mat{G}_{ij} = \mathrm{CONCAT}(\tilde{\Mat{S}}_{ij}, \bar{\Mat{S}}_{i}).
\end{equation}
Given the hybrid representation  $\Mat{G}_{ij}$, the relation generator produces a relation score as: ${w}_{ij} = f_{\phi}(\Mat{G}_{ij})$. This score will serve as the adaptive factor for the APP2S distance metric in Eq.~\eqref{eq:adaptive}.
Note that the hybrid representation for the whole support and a support-class set are denoted by
$\mathcal{G}=\{\mathcal{G}_1,\ldots,\mathcal{G}_N\}$ and $\mathcal{G}_{i}=\{\Mat{G}_{i1},\ldots,\Mat{G}_{iK}\}$, respectively.
Algorithm \ref{alg: Online Adaptation} summarizes the process of training our APP2S metric for FSL.

\begin{remark}\label{remark}
The point to set distance defined by Eq.~\eqref{eq:adaptive} is different from that in MatchingNet~\cite{vinyals2016matching}. MatchingNet formulates all the samples in the support set as a set. In contrast, we treat the samples in a class as the set, which makes our adaptive point to set distance fully contextual aware of the whole support set (by the set-signature) and encodes the distribution of each class.
\end{remark}

\begin{figure}[!h]
\begin{center}
   \scalebox{1}{
   \includegraphics[width=0.9\linewidth]{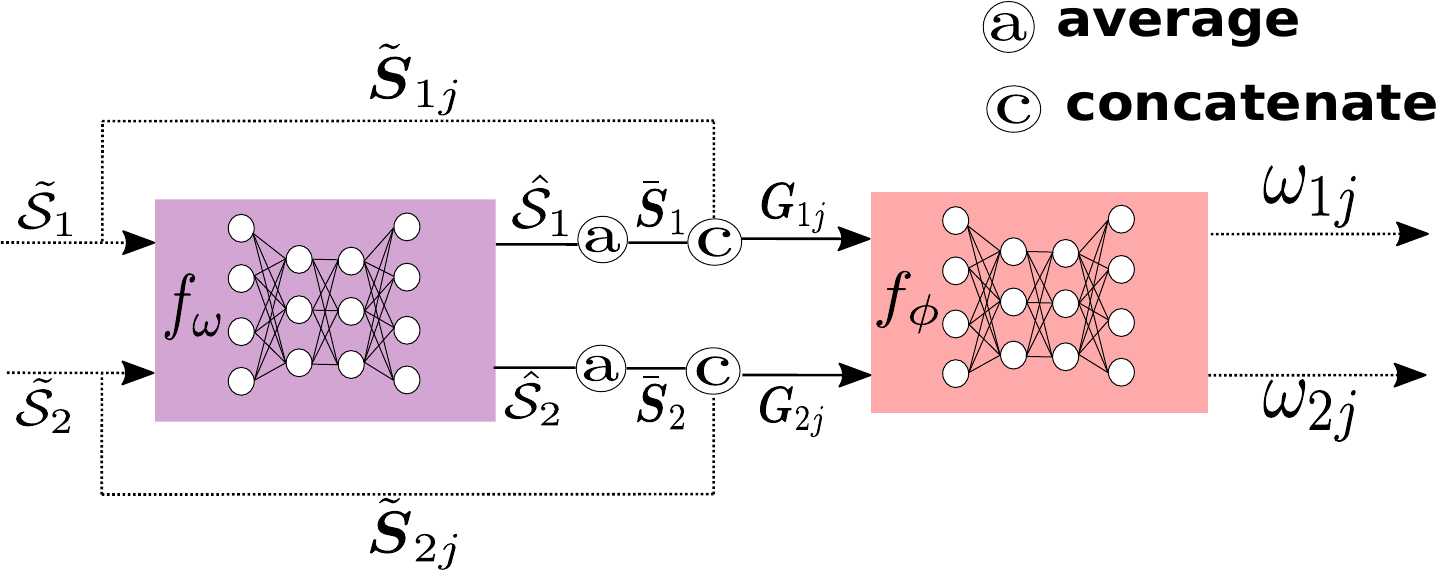}
}
\end{center}
   \caption{The information flow of the signature generator and the relation generator. The signature generator $f_{\omega}$ receives the projected support set $\{\tilde{\mathcal{S}}_1, \tilde{\mathcal{S}}_2\}$ as a bag and outputs the refined representation per sample such that each element (\ie, $\tilde{\Mat{S}}_{ij}$) sees all the other elements of the support set. Then the refined representations per class/set $\hat{\mathcal{S}}_1$ and $\hat{\mathcal{S}}_2$ are averaged to obtain the signature per class/set $\bar{\Mat{S}}_1$ and $\bar{\Mat{S}}_2$. Finally, we concatenate one projected sample $\tilde{\Mat{S}}_{1j}$ and $\tilde{\Mat{S}}_{2j}$ with the corresponding class signature $\bar{\Mat{S}}_1$ and $\bar{\Mat{S}}_2$, repectively, and feed it into relation generator $f_{\phi}$ to produce the adaptive factors  $\omega_{1j}$ and $\omega_{2j}$.}
\label{fig:sig relation}
\end{figure}

\begin{algorithm}[h]
   \caption{Train network using adaptive Poincar{\'e } point to set distance}
\textbf{Input:} An episodes $\mathcal{E}$, with their associated support set $\mathcal{X}^s = \{ (\Vec{X}^s_{ij}, y^s_{i})| i = 1, \ldots, N, j = 1, \ldots, K\}$ and a query sample $\Mat{X}^q$

\textbf{Output:} The optimal parameters for $\mathcal{F}\text{,}~ f_{\omega}, f_{\zeta},~\text{and}~f_{\phi}$
\begin{algorithmic}[1]

\State Map $\mathcal{X}^s$ and $\Mat{X}^q$ into Poincar{\'e} ball 

\State Obtain the tangent support set $\tilde{\mathcal{S}}$ using Eq.~\eqref{eq:logmap} 

\State $\hat{\mathcal{S}}=f_{\omega}(\tilde{\mathcal{S}})$ \Comment{the refined support set} 

\For {$i$ in $\{1,...,N\}$}

\State $\bar{\Mat{S}}_{i} ={ \sum_{j=1}^{K}\hat{\Mat{S}}_{ij}}/{K}$ \Comment{ the set signature}

\State $\Mat{G}_{ij}= \mathrm{CONCAT}(\tilde{\Mat{S}}_{ij}, \bar{\Mat{S}}_{i})$ 

\Comment{ the hybrid representation} 

\State  $\omega_{ij}=f_{\phi}(\Mat{G}_{ij})$\Comment{ the  weight }

\State Compute point to set distance and set to set distance using Eq.~\eqref{eq:adaptive} and Eq.~\eqref{eq: s2s}

\EndFor 
\State Optimize the model using Eq.~\eqref{eq:obj}
\end{algorithmic}
\label{alg: Online Adaptation}
\end{algorithm}

\section{Related Work}
In this section, we discuss the literature on few-shot learning and highlight those that motivate this work. Generally, there are two main branches on the few-shot learning literature, optimization-based and metric-based methods.  The optimization-based methods~\cite{antoniou2019train,chen2019closer,finn2017model, flennerhag2019meta,franceschi2018bilevel, nichol2018first}, such as MAML and Reptile~\cite{finn2017model,nichol2018first}, aim to learn a set of initial model parameters that can adapt to new tasks quickly using backpropagation in the episodic regime, without severe overfitting. However, this group of methods usually adopt a bi-level optimization setting to optimize the initial parameters, which is computationally expensive during inference. 

On the other hand, our proposed method is closer to metric-based methods~\cite{simon2020adaptive,snell2017prototypical,sung2018learning,vinyals2016matching, ye2020few,zhang2020deepemd,tang2020blockmix}, which target to realize an embedding: $\mathbb{R}^M \rightarrow \mathbb{R}^D$ to represent images in semantic space equipped with an appropriate distance metric such that different categories are distinctive. Matching Network~\cite{vinyals2016matching} determines the query labels by learning a sample-wise distance along with a self-attention mechanism that produces a fully contextualized embedding over samples. Prototypical Network~\cite{snell2017prototypical} takes a step further from a sample-wise to a class-wise metric, where all the samples of a class are averaged into a prototype to represent the class in the embedding space. Relation Network~\cite{sung2018learning} and CTM~\cite{li2019finding} replace the hand-crafted metric with a network to encode the non-linear relation between the class representations and the query embedding. Ye \etal~\cite{ye2020few} propose adopting a transformer to learn the task-specific features for few-shot learning. Zhang \etal~\cite{zhang2020deepemd} adopt the Earth Mover's Distance as  a metric to compute a structural distance between representation to obtain the labels for the query images.
Simon \etal~\cite{simon2020adaptive} propose to generate a dynamic classifier via using subspace. Along this line of research, most of the previous methods utilize the global feature vectors as representations. However, several recent works have demonstrated that utilizing the local feature maps can further boost performance. Therefore, we follow these works~\cite{doersch2020crosstransformers, zhang2020deepemd, wertheimer2021few, lifchitz2019dense, li2019revisiting} to develop our model.

However, the majority of the aforementioned metric-based works employ various metrics within Euclidean space. Ganea \etal~\cite{ganea2018hyperbolic} have proved that embedding via hyperbolic geometry enjoys low distortion for hierarchical and structured data (\eg, trees) and developed the hyperbolic version of the feed-forward neural networks and recurrent neural networks~(RNN). Moreover, a recent work~\cite{khrulkov2020hyperbolic} has shown that the vision tasks can largely benefit from hyperbolic embeddings, which inspires us to further develop algorithms with hyperbolic
geometry.

\section{Experiments}

\subsection{Datasets}

In this section, we will empirically evaluate our approach across five standard benchmarks, \ie, \emph{mini}-ImageNet~\cite{ravi2016optimization}, \emph{tiered}-ImageNet~\cite{ren2018meta}, Caltech-UCSD Birds-200-2011 (CUB)~\cite{wah2011caltech}, CIFAR-FS~\cite{bertinetto2018meta} and Fewshot-CIFAR100 (FC100)~\cite{oreshkin2018tadam}. Full details of the datasets and implementation are described in the supplementary material. 
In the following, we will briefly describe our results on each dataset.





\subsection{Main Result}
We evaluate our methods for 100 epochs, and in each epoch, we sample 100 tasks (episodes) randomly from the test set, for both 5-way 1-shot and 5-way 5-shot settings. Following the standard protocol~\cite{simon2020adaptive}, we report the mean accuracy with 95$\%$ confidence interval.

\begin{table*}[h]
    \begin{center}
    \scalebox{0.82}{
    \begin{tabular}{c c| c c|c c}
                           
        \Xhline{2\arrayrulewidth}
        \multirow{2}{*}{\bf{Model}} &\multirow{2}{*}{\bf{Backbone}}
        &\multicolumn{2}{c|}{\bf{\emph{mini}-ImageNet}}       &\multicolumn{2}{c}{\bf{\emph{tiered}-ImageNet}} \\
        
        &                           &  \bf{5-way 1-shot}        
        &\bf{5-way 5-shot}          &  \bf{5-way 1-shot}  
        &\bf{5-way 5-shot}                  \\
        \toprule\bottomrule

         ProtoNet~\cite{snell2017prototypical}
         &ResNet-12
         &$60.37\pm{0.83}$          &$78.02\pm{0.57}$
         &$61.74\pm{0.77}$          &$80.00\pm{0.55}$
         \\
         MatchingNet~\cite{vinyals2016matching}
         &ResNet-12
         &$63.08\pm{0.80}$          &$75.99\pm{0.60}$
         &$68.50\pm{0.92}$          &$80.60\pm{0.71}$
         \\
         MetaOptNet~\cite{lee2019meta}
         &ResNet-12
         &$62.64\pm{0.61}$          &$78.63\pm{0.46}$
         &$65.99\pm{0.72}$          &$81.56\pm{0.53}$
    
        \\
        Ravichandran \etal~\cite{ravichandran2019few}       &ResNet-12
        &$60.71$                     &$77.64$
        &$66.87$                     &$82.64$
        \\
        
        
        DeepEMD~\cite{zhang2020deepemd}    
        &ResNet-12
        &$65.91\pm{0.82}$              &$82.41\pm{0.56}$
        &$71.16\pm{0.87}$              &$86.03\pm{0.58}$
        \\
         P-transfer~\cite{shen2021partial}
        &ResNet-12 
        &$64.21\pm{0.77}$               &$80.38\pm{0.59}$
        &-                              &-
        
        \\
        
        GLoFA~\cite{lu2021tailoring}
        &ResNet-12
        &$66.12\pm{0.42}$              &$81.37\pm{0.33}$
        &$69.75\pm{0.33}$               &$83.58\pm{0.42}$
        
        \\
        
        DMF~\cite{xu2021learning}
        &ResNet-12
        &$\bm{67.76\pm{0.46}}$              &$82.71\pm{0.31}$
        &$71.89\pm{0.52}$               &$85.96\pm{0.35}$        
        \\
        \rowcolor{LightCyan}
         Hyperbolic ProtoNet~\cite{khrulkov2020hyperbolic}    &ResNet-12
         &$*60.65\pm{0.18}$              &$*76.13\pm{0.21}$
         &$*67.38\pm{0.14}$              &$*79.11\pm{0.22}$
        \\
        \hline
        \rowcolor{Gray}
        \bf{Ours~(APP2S)}              &ResNet-12 
        &$66.25\pm{0.20}$         &$\bm{83.42\pm{0.15}}$ 
        &$\bm{72.00}\pm{0.22}$         &$\bm{86.23\pm{0.15}}$
        \\
        \Xhline{2\arrayrulewidth}
        LwoF~\cite{gidaris2018dynamic}                       &WRN-28-10
        &$60.06\pm{0.14}$          &$76.39\pm{0.11}$
        &-                         &-
       \\   
        wDAE-GNN~\cite{gidaris2019generating}                 &WRN-28-10
        &$61.07\pm{0.15}$           &$76.75\pm{0.11}$
        &$68.18\pm{0.16}$           &$83.09\pm{0.12}$
        
        \\
        LEO~\cite{rusu2018meta}                               &WRN-28-10
        &$61.76\pm{0.08}$           &$77.59\pm{0.12}$
        &$66.33\pm{0.05}$           &$82.06\pm{0.08}$
        \\
         
        Su \etal~\cite{su2020does}                          &ResNet-18
        &-                          &$76.60\pm{0.70}$
        &-                          &$78.90\pm{0.70}$
        \\
        AFHN~\cite{li2020adversarial}                        &ResNet-18
        &$62.38\pm{0.72}$           &$78.16\pm{0.56}$
        &-                          &-
        \\
        Neg-Cosine~\cite{liu2020negative}
        &ResNet-18 
        &$62.33\pm{0.82}$        &$80.94\pm{0.59}$
        &-                       &-
        \\
        \rowcolor{LightCyan}
        Hyperbolic ProtoNet~\cite{khrulkov2020hyperbolic}   &ResNet-18
        &$*57.05\pm{0.16}$         &$*74.20\pm{0.14}$
        &$*66.20\pm{0.12}$         &$*76.50\pm{0.13}$
        \\
        \hline
         \rowcolor{Gray}
         \bf{Ours~(APP2S)}                &ResNet-18  
         &$64.82\pm{0.12}$     
         &$81.31\pm{0.22}$
         &$70.83\pm{0.15}$     
         &$84.15\pm{0.29}$
        
        \\
    \Xhline{2\arrayrulewidth}
    \end{tabular}
    }
    \end{center}
    \caption{Few-shot classification accuracy and 95 \% confidence interval on \emph{mini}-ImageNet and \emph{tiered}-ImageNet with ResNet backbones. ``*" notes the result obtained by the self-implemented network.}
    \label{tab: mini and tier}
\end{table*}

\begin{table*}[h]
    \begin{center}
    \scalebox{0.9}{
    \begin{tabular}{c c| c c | c c}
        \Xhline{2\arrayrulewidth}
         \multirow{2}{*}{\bf{Model}}       &\multirow{2}{*}{\bf{Backbone}}
          &\multicolumn{2}{c|}{\bf{CIFAR-FS}}     &\multicolumn{2}{c}{\bf{FC100}}
        \\
                                         &
         &\bf{5-way 1-shot}              &\bf{5-way 5-shot}           
         &\bf{5-way 1-shot}              &\bf{5-way 5-shot} 
        \\
      \toprule\bottomrule
        TEAM~\cite{qiao2019transductive} &ResNet-12
        &$70.40$                    &$81.30$ 
        &-                              &-
        \\
        ProtoNet~\cite{snell2017prototypical} &ResNet-12
        &$72.20\pm{0.70}$                 &$83.50\pm{0.50}$ 
        &$37.50\pm{0.60}$                 &$52.50\pm{0.60}$ 
        \\ 
        TADAM~\cite{oreshkin2018tadam}    &ResNet-12
        &-                                &-
        &$40.10\pm{0.40}$                 &$56.10\pm{0.40}$ 
      
        \\
        DeepEMD~\cite{zhang2020deepemd} &ResNet-12
        &-               &- 
        &$46.47\pm{0.78}$               &$63.22\pm{71}$ 
          \\
        \rowcolor{LightCyan}
        Hyperbolic ProtoNet~\cite{khrulkov2020hyperbolic} &ResNet-12
        &*$70.27\pm{0.22}$               &*$80.98\pm{0.16}$ 
        &*$36.04\pm{0.18}$               &*$51.60\pm{0.18}$

        \\
        \hline
        \rowcolor{Gray}
        \bf{Ours~(APP2S)}             &ResNet-12
        & $\bm{73.12\pm{0.22}}$     & $\bm{85.69\pm{0.16}}$  
        & $\bm{47.64\pm{0.21}}$     &$\bm{63.56\pm{0.22}}$
      \\
    \Xhline{2\arrayrulewidth}
    \end{tabular}
    }
    \end{center}
    \caption{Few-shot classification accuracy and 95 \% confidence interval on CIFAR-FS and FC100 with ResNet-12 backbones. ``*" notes the result obtained by the self-implemented network.}
    \label{tab: cifar}
\end{table*}

\begin{table}[h]
    \begin{center}
    \scalebox{0.7}{
    \begin{tabular}{c | c c}
        \Xhline{2\arrayrulewidth}
         \bf{Model}                     
         &\bf{5-way 1-shot}              &\bf{5-way 5-shot}     
        \\
      \toprule\bottomrule
        MAML~\cite{finn2017model}                       
        &$69.96\pm{1.01}$          &$82.70\pm{0.65}$
        \\
        RelationNet~\cite{sung2018learning}             
        &$67.59\pm{1.02}$          &$82.75\pm{0.58}$
        \\
        Chen \etal~\cite{chen2019closer}                
        &$67.02$                   &$83.58$
        \\
        MatchingNet~\cite{vinyals2016matching}          
        &$72.36\pm{0.90}$          &$83.64\pm{0.60}$
        
         \\
       
        SimpleShot~\cite{wang2019simpleshot}                              
        &$70.28$                   &$86.37$
        \\
       
        ProtoNet~\cite{snell2017prototypical}           
        &$71.88\pm{0.91}$          &$87.42\pm{0.48}$ 
        \\
        
        DeepEMD~$^\clubsuit$~\cite{zhang2020deepemd}             
        &$76.65\pm{0.83}$          &$88.69\pm{0.50}$
        \\
        
        P-transfer~$^\clubsuit$~\cite{shen2021partial}
        &$73.88\pm{0.87}$               &$87.81\pm{0.48}$
        
        \\
        \rowcolor{LightCyan}
        Hyperbolic ProtoNet~\cite{khrulkov2020hyperbolic}   
         &$*73.70\pm{0.22}$        &$*85.55\pm{0.13}$

        \\
        
        \hline
        \rowcolor{Gray}
        \bf{Ours~(APP2S)}           
        & $\bm{77.64\pm{0.19}}$     & $\bm{90.43\pm{0.18}}$  
      \\
    \Xhline{2\arrayrulewidth}
    \end{tabular}
    }
    \end{center}
    \caption{Few-shot classification accuracy and 95 \% confidence interval on CUB. ``*" notes the result obtained by the self-implemented network. ``$\clubsuit$" denotes the method using ResNet-12 as the backbone, otherwise ResNet-18.}
    \label{tab: cub}
\end{table}

\noindent{\bf{\emph{mini}-ImageNet}}. As shown in Table~\ref{tab: mini and tier}, we evaluate our model using ResNet-12 and ResNet-18  as the backbones on \emph{mini}-ImageNet. Between them, ResNet-12 produces the best results. In addition, our model also outperforms recent state-of-the-art models in most of the cases. Interestingly, our model further outperforms hyperbolic ProtoNet by 7.77\% and 7.11\% for 5-way 1-shot and 5-way 5-shot with ResNet-18, respectively. With ResNet-12, we outperform the hyperbolic ProtoNet by 5.60\% and 7.29\%  for 5-way 1-shot and 5-way 5-shot, respectively.

\noindent{\bf{\emph{tiered}-ImageNet}}. We further evaluate our model on \emph{tiered}-ImageNet with ResNet backbones. The results in Table \ref{tab: mini and tier} indicate that with ResNet-12, our model outperforms the hyperbolic ProtoNet by 4.62\% and 7.12\% for 5-way 1-shot and 5-way 5-shot, respectively, and achieves state-of-the-art results for inductive few-shot learning.

\noindent{\bf{CIFAR-FS} and \bf{FC100}}. As the results in Table \ref{tab: cifar} suggested, our model also achieves comparable performance with the relevant state-of-the-state methods on this dataset, with ResNet-12 backbone, which vividly shows the superiority of our method.

\noindent{\bf{CUB.}} We use ResNet-18 as our backbone to evaluate our method on the CUB dataset. Table~\ref{tab: cub} shows that our model improves the performance over baseline by 3.94\% and 4.88\% for 5-way 1-shot and 5-way 5-shot settings, respectively. Besides, our model achieves 77.64\% and 90.43\% for 5-way 1-shot and 5-way 5-shot settings on this dataset, which outperforms state-of-the-art models (\ie, DeepEMD~\cite{zhang2020deepemd} and P-transfer~\cite{shen2021partial}) and achieve competitive performance on this dataset.

\subsection{Robustness to Outliers} \label{Robutness analysis}

\begin{figure}[!h]
\centering
	\subfigure[]{\includegraphics[width=0.48\linewidth]{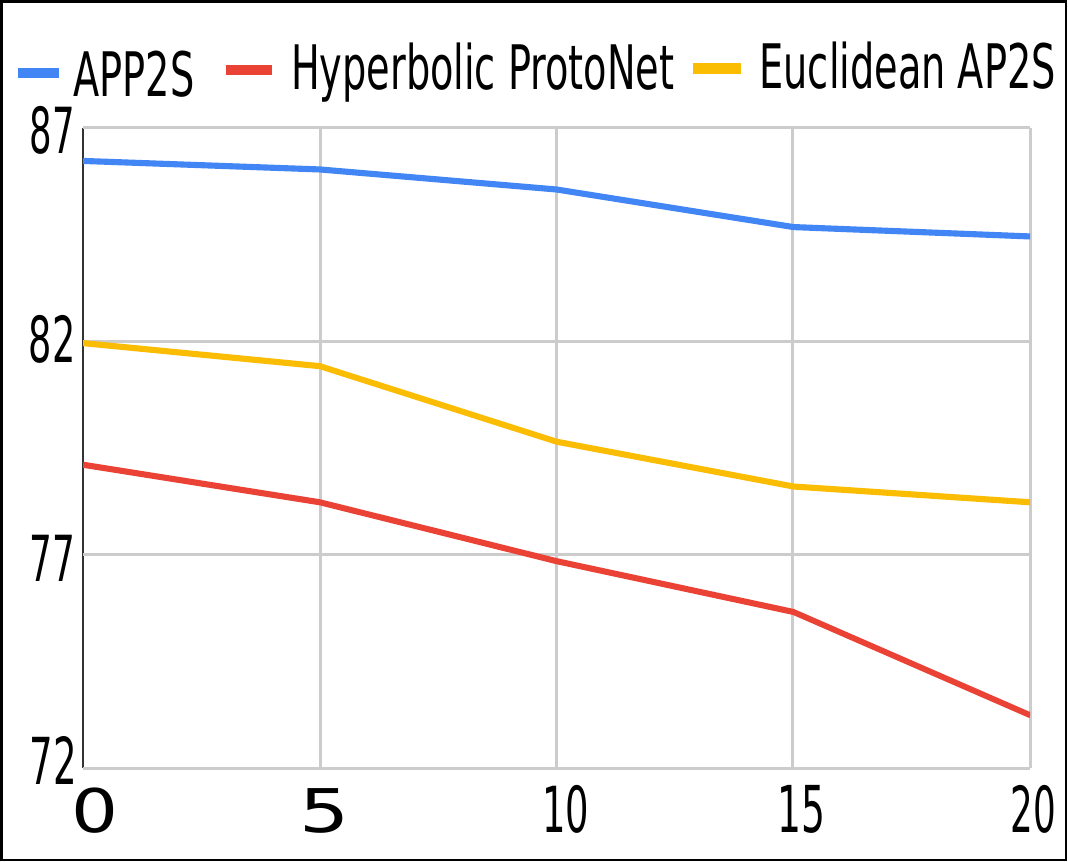}\label{fig:outa}}%
	\hfil
	\subfigure[]{\includegraphics[width=0.48\linewidth]{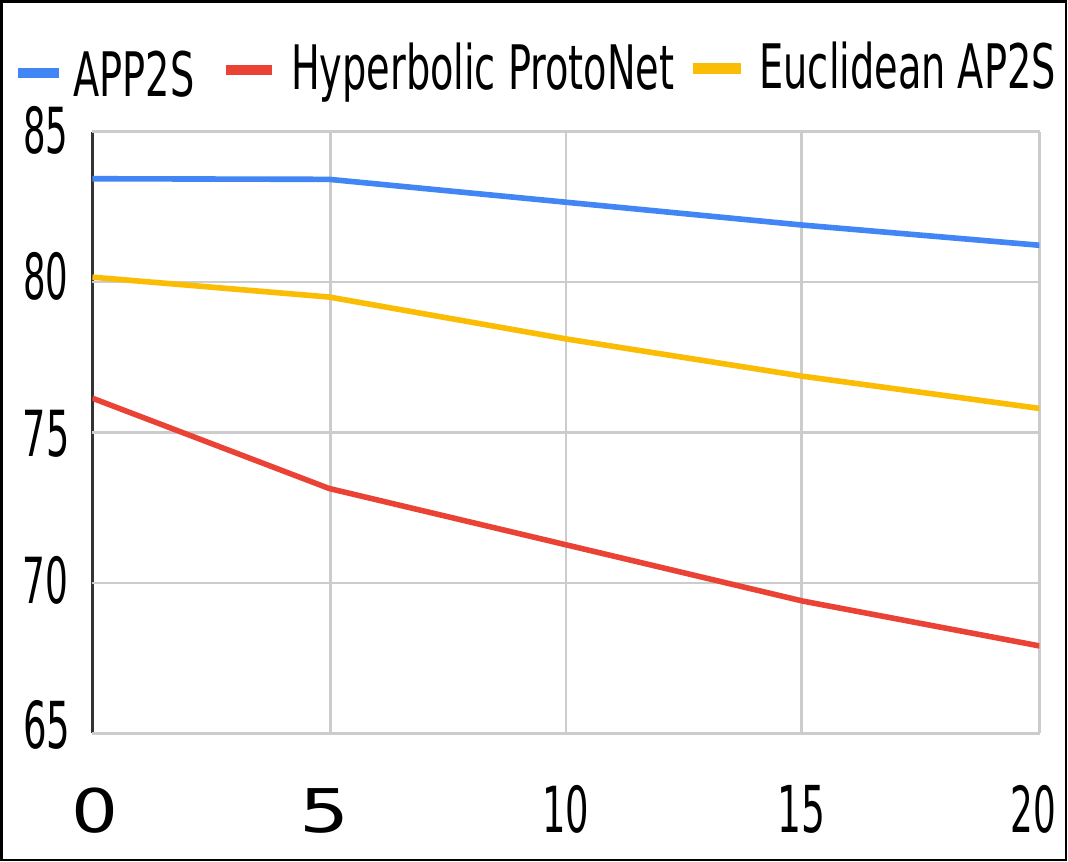}\label{fig:outb}}%
	\hfil
	\caption{Robustness analysis. Horizontal axis: The number of outliers. Vertical axis: Accuracy. (a): The performance vs. the number of outliers on \emph{tiered}-ImageNet. (b): The performance vs. the number of outliers on \emph{mini}-ImageNet. 
	}\label{fig:compare_1}
\end{figure}

To further validate the robustness of our method, we conduct experiments in the presence of outliers in the form of mislabelled images. In the first study, we add a various number of outliers (\eg, 1, 2, 3, 4), whose classes are disjoint to the support-class, to each class of the support set. We performed this study with ResNet-12 backbone on the 5-way 5-shot setting on \emph{tiered}-ImageNet. Fig.~\ref{fig:outa} shows that the performances of hyperbolic ProtoNet degrade remarkably. On the contrary, both our APP2S and Euclidean AP2S are robust to outliers, which shows the superiority of our adaptive metric. Comparing to Euclidean AP2S, APP2S is even more robust (see the slope of Fig.~\ref{fig:outa}) and performs consistently even in the presence of 20 outliers. This suggests that integrating our proposed adaptive metric and hyperbolic geometry can further bring robustness to our framework. In the second study (shown in Fig.~\ref{fig:outb}), we conduct the same experiments on \emph{mini}-ImageNet. The results show a similar trend as the previous one, which further proves the effectiveness of our proposed method.

\subsection{Ablation Study}
\label{Ablation Study}
We further conduct the ablation study to verify the effectiveness of each component in our method on the \emph{tiered}-ImageNet dataset using the ResNet-12 backbone. 

\noindent{\bf{Experiments Set-Up}}. For setting (ii) in Table~\ref{tab: ablation study}, we disable the relation module $f_{\phi}$ and signature generator $f_{\omega}$. The P2S distance can be obtained by Eq.~\eqref{eq:adaptive} and Eq.~\eqref{eq: s2s} with equal weights (\ie, $1$). Moreover, we enable the relation generator $f_{\phi}$ but not the signature generator in setting (iii). We use the class prototype instead of the signature for this experiment. We enable both $f_{\phi}$ and $f_{\omega}$ and use the Euclidean distances for setting (iv). In the end, we enable the Poincar{\'e} ball but disable the $f_{\zeta}$ for setting (v). In terms of implementation of (v), the backbone is designed to output a feature vector instead of a feature map, such that the P2S distance can be directly computed by Eq.~\eqref{eq:distance} and Eq.~\eqref{eq:adaptive}.

\noindent{\bf{Effectiveness of Point to Set Distance}}. In this experiment, we first evaluate the effectiveness of the P2S distance by comparing to its point to point (P2P) distance counterpart (\ie, hyperbolic ProtoNet). From Table~\ref{tab: ablation study}, we could observe that the P2S distance can learn a more discriminative embedding space than P2P distance (\ie, (i) vs. (ii)), and the adaptive P2S can further bring performance gain to our application (\ie, (ii) vs. (iii)). This observation shows the potential of our P2S distance setting in the FSL task.

\noindent{\bf{Effectiveness of Signature Generator}}.
We further evaluate another essential component in our work, \ie, the signature generator, which refines the entire support set and produces a signature per class. As shown in Table~\ref{tab: ablation study} (\ie, (iii) and (vi)), we could observe that our method benefits from the signature generator, which shows that the signature of each class could help to generate an informative weight for individual feature map within the same class. 

\noindent{\bf{Effectiveness of Hyperbolic Geometry}}. We also implement our model in the Euclidean space to verify the effectiveness of our method. The row (iv) and (vi) in Table \ref{tab: ablation study} vividly show that the representation in the Poincar\'e ball has a richer embedding than that in Euclidean spaces.

\noindent{\bf{Effectiveness of Set to Set Distance}}.
The comparison between (v) and (vi) shows that our set to set distance generator associated with the feature map outputs richer information than using a feature vector to directly compute the APP2S.

\begin{table}[H]
    \begin{center}
    \scalebox{0.72}{
    \begin{tabular}{c| c |  c|c|c|c| c| c  }
                           
        \Xhline{2\arrayrulewidth}
        
          \multirow{2}{*}{\bf{ID}}
          &\multirow{2}{*}{\bf{Model}} 
          &\multirow{2}{*}{$\mathbb{{D}}^n_c$}
          &\multirow{2}{*}{P2S}
          &\multirow{2}{*}{${f}_{\phi}$}
          &\multirow{2}{*}{${f}_{\omega}$}
          &\multirow{2}{*}{${f}_{\zeta}$}
          &{\bf{\emph{tiered}-ImageNet}}     
        \\
          &&&&&&                               
          &\bf{5-way 5-shot}          
        \\
       \toprule\bottomrule
        (i)
        &Hyperbolic ProtoNet
        & \checkmark
        &
        &
        &
        & 
        &$79.11\pm{0.22}$               
       \\
     (ii)
     &Hyperbolic P2S w/o~$f_{\phi}$
     &\checkmark
     &\checkmark
     &
     &
     &\checkmark
     &$83.14\pm{0.17}$                
        \\
       (iii)
       &Hyperbolic P2S w/~$f_{\phi}$
       &\checkmark 
       &\checkmark
       &\checkmark
       &
       &\checkmark
      &$84.88\pm{0.17}$   
       \\
    (iv)
    &Euclidean AP2S
    &
    &\checkmark
    &\checkmark
    &\checkmark
    &\checkmark
    &$81.96\pm{0.18}$
    \\
    (v)
    & APP2S w/o $f_{\zeta}$
    &\checkmark
    &\checkmark
    &\checkmark
    &\checkmark
    &
    &$84.12\pm{0.13}$
    \\
    \hline
     \rowcolor{Gray}
     (vi)
     &\bf{APP2S}
     &\checkmark
     &\checkmark
     &\checkmark
     &\checkmark
     &\checkmark
     &$\bm{86.23\pm{0.15}}$   
      \\
    \Xhline{2\arrayrulewidth}
    \end{tabular}
    }
    \end{center}
    \caption{The ablation study of our model, we start from the hyperbolic ProtoNet~\cite{khrulkov2020hyperbolic} towards APP2S.}
    \label{tab: ablation study}
\end{table}

\section{Conclusion}
In this paper, we propose a novel adaptive Poincar{\'e} point to set (APP2S) distance metric for the few-shot learning, which can adapt depending on the samples at hands. Empirically, we showed that this approach is  expressive with both hyperbolic geometry and Euclidean counterpart.
Our model improves the performances over baseline models and achieves competing results on five standard FSL benchmarks.

\section{Supplementary Material}
In this supplementary material, we provide an additional description of operations in Poincar{\'e} Ball and details of the each public few-shot learning benchmark we used. Furthermore, we conduct additional experiments, including ablation studies on the effect of the curvature $c$, global feature vector implementation and parameter and time complexity analysis to analyze the our model. Finally, we provide the details of the implementation of our model and extra visualizations and discussion of APP2S.

\subsection{Hyperbolic Operations}
{\bf{\noindent{Exponential Map}}}. The exponential map defines a function from $T_{\Vec{x}}\mathbb{D}_c^n   \to \mathbb{D}^n_c $, which maps Euclidean vectors to the hyperbolic space. Formally, it is defined as: 

\begin{equation}
    \label{eq:expmap}
    \Vec{\Omega}^c_\Vec{x}(\Vec{v})=\Vec{x}\oplus_c(\text{tanh}(\sqrt{c}\frac{\lambda^c_{\Vec{x}}\|\Vec{v}\|}{2})\frac{\Vec{v}}{\sqrt{c}\|\Vec{v}\|}).
\end{equation}

The exponential map and logarithmic map (introduced in the main paper) have simpler forms when $\Vec{x}=0$:
\begin{equation}
    \label{eq:simpler}
    \Vec{\Omega}^c_{\bm{0}}(\Vec{v})=\text{tanh}(\sqrt{c}\|\Vec{v}\|)\frac{\Vec{v}}{\sqrt{c}\|\Vec{v}\|} ,
\end{equation}

\begin{equation}
    \label{eq:simpler1}
   \Vec{\pi}^c_{\bm{0}}(\Vec{y})=\text{arctanh}(\sqrt{c}\|\Vec{y}\|)\frac{\Vec{y}}{\sqrt{c}\|\Vec{y}\|}.
\end{equation}

\noindent{\bf{Parallel Transport}}. Parallel transport provides a way to move tangent vectors along geodesics $P_{\Vec{x}\to\Vec{y}}: T_\Vec{x}\mathcal{M}  \xrightarrow{}T_\Vec{y}\mathcal{M}$ and defines a canonical way to connect tangent spaces. For further details of hyperbolic space and geometry, please refer to the thesis~\cite{ganea2019non}.

\subsection{Set to Set Distance}
Set to set distance has been widely adopted in computer vision tasks~\cite{Fang_2021_WACV, Conci2018DistanceBS,huttenlocher1993comparing}. In this section, we discuss the well-known Hausdorff distance. There are two variants of Hausdorff distance, including one-sided Hausdorff distance and bidirectional Hausdorff distance. 
The one-sided Hausdorff distance between set $A=\{\Vec{a}_1,\Vec{a}_2,\ldots,\Vec{a}_n\}$ and set $B=\{\Vec{b}_1,\Vec{b}_2,\ldots,\Vec{b}_n\}$ can be defined as:
\begin{equation}\label{eq:one-sided Hausdorff}
{\mathrm{d_{Hau}^{o}}}(A,B)=\max_{\Vec{a}\in A}\min_{\Vec{b}\in B}\mathrm{d}(\Vec{a},\Vec{b}),
\end{equation}
and the bidirectional Hausdorff distance can be defined as:
\begin{equation}\label{eq:bi Hausdorff}
\mathrm{d_{Hau}^{bi}}(A,B)=\max({\mathrm{d_{Hau}^{o}}}(A,B), {\mathrm{d_{Hau}^{o}}}(B,A)).
\end{equation}


\subsection{Datasets}
\noindent{\bf{\emph{mini}-ImageNet}}. The \emph{mini}-ImageNet is a subset of ImageNet~\cite{deng2009imagenet}. The size of images in \emph{mini}-ImageNet is fixed to 84 $\times$ 84. It has 100 classes, with each having 600 samples. We adopt the standard setting form~\cite{ravi2016optimization} to split the dataset into 64, 16, and 20 classes for training, validation, and testing.

\noindent{\bf{\emph{tiered}-ImageNet}}. Like \emph{mini}-ImageNet, \emph{tiered}-ImageNet~\cite{ren2018meta} is also sampled from ImageNet, while it has more classes than the \emph{mini}-ImageNet. This dataset is split into 351 classes from 20 categories, 97 classes from 6 categories, and 160 classes from 8 different categories for training, validation, and testing. 

\noindent{\bf{CUB}}. The CUB dataset~\cite{wah2011caltech} consists of 11,788 images from 200 different species of birds. Following the standard split~\cite{liu2020negative}, the CUB dataset is divided into 100 species for training, 50 species for validation, and another 50 species for testing.      

\noindent\textbf{CIFAR-FS} and \textbf{FC100}. Both CIFAR-FS~\cite{bertinetto2018meta} and FC100~\cite{oreshkin2018tadam} are modified from the CIFAR-100 dataset containing 100 classes, with 600 samples per class. The CIFAR-FS is split into 64, 16, and 20 classes for training, validation, and testing, respectively. While the FC100 dataset is split into 60, 20, and 20 classes for training, validation, and testing, respectively.

\subsection{Additional Experiments}
\noindent{\bf{Conv-4 Backbone.}}
 We also employ the simple 4-convolutional network (Conv-4) to evaluate our method on \emph{mini-}ImageNet comparing with some early works. The Table~\ref{tab: conv} summarizes our results.
 
 \begin{table}[h]
    \begin{center}
    \scalebox{0.7}{
    \begin{tabular}{c | c c}
                           
        \Xhline{2\arrayrulewidth}
        \bf{Model}                
        &  \bf{5-way 1-shot}   &  \bf{5-way 5-shot} \\
        \toprule\bottomrule

       MatchingNet~\cite{vinyals2016matching}
       &$43.56\pm{0.84}$          &$55.31\pm{0.73}$\\    
        MAML~\cite{finn2017model}
        &$48.70\pm{1.84}$         &$63.11\pm{0.92}$\\  
        
        RelationNet~\cite{sung2018learning}
        &$50.44\pm{0.82}$        &$65.32\pm{0.70}$ \\  
        
        R2-D2~\cite{bertinetto2018meta} 
        &$48.70\pm{0.60}$        &$65.50\pm{0.60}$ \\  
        
        Reptile~\cite{nichol2018first} 
        &$49.97\pm{0.32}$        &$65.99\pm{0.58}$\\  
        
        ProtoNet~\cite{snell2017prototypical}
        &$49.42\pm{0.78}$        &$68.20\pm{0.66}$ \\  
        
        Neg-Cosine~\cite{liu2020negative}
        &$52.84\pm{0.76}$        &$70.41\pm{0.66}$
                            \\
        \rowcolor{LightCyan}
        Hyperbolic ProtoNet~\cite{khrulkov2020hyperbolic}  
        &$54.43\pm{0.20}$        &$72.67\pm{0.15}$ \\  
        \hline
        \rowcolor{Gray}
        \bf{Ours~(APP2S)} 
         &$\bm{55.73\pm{0.20}}$  &$\bm{72.86\pm{0.22}}$ 
        \\
    \Xhline{2\arrayrulewidth}
    \end{tabular}
    }
    \end{center}
   
    \caption{Few-shot classification accuracy and 95 \% confidence interval on \emph{mini}-ImageNet with Conv-4 Backbone.}
    \label{tab: conv}
\end{table}
 
\noindent{\textbf{Comparison with DN4 and FEAT.}} The comparison of our APP2S, DN4, and FEAT on \emph{mini}-ImageNet and \emph{tiered}-ImageNet with Conv-4 and ResNet-12 backbones is summarized in Table~\ref{tab:com}.
\begin{table}[h]
    \begin{center}
    \scalebox{0.75}{
    \begin{tabular}{c|c|c c|c c}
        \Xhline{2\arrayrulewidth}
        \multirow{2}{*}{\textbf{Model}}
        &\multirow{2}{*}{\textbf{Backbone}}
        &\multicolumn{2}{c|}{\textbf{\emph{mini}-ImageNet}}
        &\multicolumn{2}{c}{\textbf{\emph{tiered}-ImageNet}}\\
        &
        &1-shot &5-shot
        &1-shot &5-shot
        
        \\
        \toprule\bottomrule
        {DN4~\cite{li2019revisiting}}
        &Conv4
        & $51.24$  & ${71.02}$  & -  &-
        \\
        {FEAT~\cite{ye2020few}} 
        &Conv-4
        & $55.15$  & ${71.13}$ &-   &- 
        \\
        \rowcolor{LightCyan}
        {Ours} 
        &Conv-4
        & $\bm{55.73}$  & $\bm{72.86}$ &-    &- 
        \\
        \hline
        {FEAT} 
        &ResNet-12
        & $\bm{66.78}$  & $82.05$ &$70.80$   &$84.79$ 
        \\
        \rowcolor{LightCyan}
        {Ours} 
        &ResNet-12
        & $66.25$  & $\bm{83.42}$  &$\bm{72.00}$    &$\bm{86.23}$ 
        \\

        \Xhline{2\arrayrulewidth}
    \end{tabular}
    }
    \end{center}
    \caption{The comparison of our model, DN4 and FEAT on the \emph{mini}-ImageNet and \emph{tiered}-ImageNet with Conv-4 and ResNet-12 backbones.}
    \label{tab:com}
\end{table}

\noindent{\bf{The Curvature of Poincar{\'e} ball.}}
The curvature of the Poincar{\'e} ball is an important parameter, which determines the radius of the Poincar{\'e} ball. We conduct experiments with different values of $c$ on \emph{tiered}-ImageNet. The results are summarized into Table~\ref{tab: c}. As the results suggested, our model is not very sensitive to $c$. However, with a larger $c$ value, the performance is slightly better.

\begin{table}[H]
    \begin{center}
    \scalebox{0.8}{
    \begin{tabular}{c |c c c c c c}
                           
        \Xhline{2\arrayrulewidth}
    
          \bf{Model}
          & $\bm{0.7}$
          & $\bm{0.5}$
          & $\bm{0.1}$
          & $\bm{0.05}$
          & $\bm{0.01}$
          & $\bm{0.01}$
        \\
       \toprule\bottomrule
        APP2S                
        &$\bm{86.23}$
        &$85.92$                
        &$85.21$   
        &$85.18$
        &$84.43$
        &$84.19$
      \\

    \Xhline{2\arrayrulewidth}
    \end{tabular}
    }
    \end{center}
    \caption{The influence from the curvature of Poincar{\'e} ball on the performance, given \emph{tiered}-ImageNet and ResNet-12 backbone on 5-way 5-shot setting.}
    \vspace{0.5em}
    \label{tab: c}
\end{table}

\noindent{\bf{1-shot case}.} To fully leverage the capability of APP2S for 1-shot setting, we require more than one sample in the set. Therefore, we followed the practice in~\cite{simon2020adaptive} to augment the support images by flipping. To have a fair comparison, we also applied augmentation to our baseline model (\ie, hyperbolic ProtoNet~\cite{khrulkov2020hyperbolic}) on both \emph{mini}-ImageNet and \emph{tiered}-ImageNet, given ResNet-12 backbone. Table~\ref{tab: aug} shows that the image augmentation does not boost the performance of the baseline model significantly.

\begin{table}[H]
    \begin{center}
    \scalebox{0.78}{
    \begin{tabular}{c |c c| c c }
                           
        \Xhline{2\arrayrulewidth}
    
          \multirow{2}{*}{\bf{Model}}
          &\multicolumn{2}{c|}{\bf{\emph{mini}-ImageNet}} 
          &\multicolumn{2}{c}{\bf{\emph{tiered}-ImageNet}}
        \\
        &\bf{w/o Aug.}
        &\bf{w/ Aug.}
        &\bf{w/o Aug.}
        &\bf{w/ Aug.}
        
        \\
       \toprule\bottomrule
        hyperbolic ProtoNet
        &$*60.65$
        &$*59.97$       
        &$*67.38$   
        &$*67.84$

        \\
        APP2S                
        &-
        &$66.25$                
        & -  
        &$72.00$
      \\
    \Xhline{2\arrayrulewidth}
    \end{tabular}
    }
    \end{center}
    \caption{The accuracy without and with augmentation for 1-shot setting. ``*" notes the results obtained by self-implemented network.}
    \vspace{0.5em}
    \label{tab: aug}
\end{table}

\noindent{\bf{any-shot setting}.} We follow the any-way \& any-shot setting introduced in~\cite{lee2019learning} to further validate the efficacy of our algorithm. We use a variant of our final model (APP2S without $f_{\zeta}$) to perform this experiments due to less computation requirement on this experiment setting.
The results are shown in Table~\ref{tab:any}. 

\begin{table}[!h]
    \begin{center}
    \scalebox{0.6}{
    \begin{tabular}{c | c  c | c  c }
        \hline
          \multirow{2}{*}{Model}
         & \multicolumn{2}{c|}{{\emph{tiered}-ImageNet}}
         & \multicolumn{2}{c}{{\emph{mini}-ImageNet}}
         \\
         &Conv-4 &ResNet-12  &Conv-4 &ResNet-12\\
    \hline
        ProtoNet
        &-&-&$51.02\pm{0.30}$ &-
        \\
        L2G ProtoNet\cite{lee2019learning}
        &-&-&$53.00\pm{0.28}$&-
        \\
        \hline
        \textbf{Ours} &$\bm{66.23\pm{0.15}}$ &$\bm{75.52\pm{0.19}}$ &$\bm{58.62\pm{0.16}}$ &$\bm{70.01\pm{0.17}}$ 
        \\
     \Xhline{2\arrayrulewidth}
    \end{tabular}
    }
    \end{center}
    \vspace{-1em}
    \caption{The results of our model under any-way \& any-shot setting compared to ProtoNet and L2G PrptoNet.}
    
    \label{tab:any}
\end{table}

\noindent{\bf{Using Global Feature Vectors.}} We performed extra experiments using global feature vectors in our method. The  table below shows that our method, even with global feature vectors, outperforms the Hyperbolic ProtoNet significantly, and the local features can further boost our methods. Note that we use ResNet-18 backbone for CUB dataset and ResNet-12 for the rest.

\begin{table}[h]
    \begin{center}
    \scalebox{0.62}{
    \begin{tabular}{c|c|c|c}
        \hline
        {Dataset}&Hyperbolic ProtoNet
        &Ours w/o global feature &Ours w/ local feature\\
        \hline
        {\emph{mini}-ImgeNet} & $76.13\pm{0.21}$  & ${80.13\pm{0.11}}$& 
        ${83.42\pm{0.15}}$\\
        \hline
        {\emph{tiered}-ImageNet} &$79.11\pm{0.22}$ & ${84.12\pm{0.13}}$ &
        ${86.23\pm{0.15}}$
        \\
        \hline
        {CUB} &$85.55\pm{0.13}$ & ${90.24\pm{0.15}}$&
        ${90.43\pm{0.18}}$
        \\
        \hline
        {CIFAR-FS} & ${80.98\pm{0.16}}$ &  ${84.08\pm{0.16}}$&
        ${85.69\pm{0.16}}$
        \\
        \hline
    \end{tabular}
    }
    \end{center}
    \caption{The results of our model using the global feature vectors compared to Hyperbolic ProtoNet.}
    \label{tab: global and local}
\end{table}

\noindent{\bf{Parameter and time complexity analysis.}} 
Comparing to Hyperbolic ProtoNet, we have $3$ extra modules, including  $f_{\omega}$, $f_{\phi}$ and $f_{\zeta}$ to realize the adaptive distances. We summarize the parameter numbers (PNs) and FLOPs for each module and the backbone network. We can find that the PNs and FLOPs of our module are acceptable as compared with the backbone network. We also compare the time complexity to the SOTA method, \ie, DeepEMD, given that both methods are using local feature maps. The FPS value of our method is $83$, as compared to $0.6$ of DeepEMD under the 5-shot setting, clearly showing that our method runs faster than DeepEMD. Note that both models are tested on a single Nvidia Quadro-GV100 graphic card. 

\begin{table}[h]
    \begin{center}
    \scalebox{1}{
    \begin{tabular}{c|c|c|c|c}
        \hline
        complexity metrics
        &$f_{\omega}$   &$f_{\phi}$   &$f_{\zeta}$    &$f_{\theta}$\\
        \hline
        
        PNs ($\times10^6$)     
        &$1.64$         &$0.74$       &$0.02$         &$\bm{12.42}$\\
        \hline
        
        FLOPs ($\times10^9$)
        &$2.01$         &$0.17$       &$0.0008$       &$\bm{6.98}$\\
        \hline

    \end{tabular}
        }
    \end{center}
    \caption{The Parameter and time complexity analysis.}
    \label{tab: time complexity}
\end{table}

\subsection{Implementation Details}
\noindent{\bf{Network and Optimizer}}. We mainly use ResNet~\cite{he2016deep}, including ResNet-12 and ResNet-18, as our backbones across all datasets. We also employ the simple 4-convolutional network (Conv-4) to evaluate our method comparing with some early works. The size of the input image is fixed to 84 $\times$ 84. We use Adam~\cite{kingma2014adam} and SGD~\cite{ye2020few} for Conv-4 and ResNet backbones, respectively. In the SGD optimizer, we adopt the L2 regularizer with 0.0005 weight decay coefficient. In the ResNet-12 backbones, we disable the average pooling and remove the last fully connected (FC) layer, such that the networks generate the feature map with size of $640 \times 5 \times 5$. For ResNet-18, we set the average pooling layer to generate the feature map with the size of $512\times 5  \times 5$. Note that we set $c$ (the curvature of the Poincar{\'e} ball) to 0.7 and 0.5 for 5-way 5-shot setting and for 5-way-1-shot setting, respectively, with ResNet backbones, across all datasets. While for Conv-4, we set $c$ to 0.4 for both 5-way 5-shot and 5way 1-shot settings across all datasets.   


\noindent{\bf{Training}}. Following the excellent practice in state-of-the-art methods~\cite{ye2020few,zhang2020deepemd,simon2020adaptive}, network training has two stages, \ie, pre-training stage and meta-learning stage. In the pre-training process, the backbone network followed by a FC layer is trained on all training classes with the standard classification task. The network with the highest validation accuracy is selected as the pre-trained backbone for the next training stage. In the meta-learning stage, we also follow the standard training protocol, where the network is trained for 200 epochs, and each epoch samples 100 tasks randomly. In order to create the set for 5 way 1-shot setting, we follow the previous practice in~\cite{simon2020adaptive}, which augments the image per class by horizontal flipping.

\noindent{\bf{Signature Generator}}. For the signature generator, we choose Transformer encoder as the set refinement function $f_{\omega}$ as it performs contextualization over the whole support set with permutation invariant property. Note that the Transformer is implemented with single-head self-attention because more heads do not boost the performance but require more computational power for our model by experiments. Moreover, We follow the implementation of~\cite{carion2020end} to provide spatial encoding along with flattened feature map into the transformer. 

\noindent{\bf{Relation Generator}}. We implemented the relation generator using a simple two-layer CNN followed by a flatten operation in the end. In the first layer, the linear transformation is followed by the batch normalization and activation. The second layer uses the sigmoid function to bound the output. Finally, a softmax layer is implemented to convert the output into a probability distribution. The structure of the relation generator can be summarized into Table~\ref{tab:r_structure}.

\begin{table}[H]
    \begin{center}
    \scalebox{1}{
    \begin{tabular}{c|c| c }
                           
        \Xhline{2\arrayrulewidth}
    
            \bf{layer name}
          & \bf{output size}
          & \bf{operation parameter}
         
        \\
       \toprule\bottomrule
        conv1     & $3 \times 3$ & $3 \times 3$, 64, stride 1                 
        \\
       \hline 
       batch norm & $3 \times 3$ &64
        \\
       \hline 
       relu       &$3 \times 3$  &-
        \\
       \hline 
       dropout    &$3 \times 3$  & $p=0.5$
        \\
       \hline
       conv2      &$1 \times 1$  &$3 \times 3$, 1, stride 1                 
        \\
       \hline 
       batch norm &$1 \times 1$  &1
        \\
       \hline 
       sigmoid    &$1 \times 1$  &-
      \\
    \Xhline{2\arrayrulewidth}
    \end{tabular}
    }
    \end{center}
    \caption{The structure of the Relation Generator.}
    \vspace{0.5em}
    \label{tab:r_structure}
\end{table}

\noindent{\bf{Set to Set Distance Generator}}. We simply implement a two layer MLP (\ie, $625\rightarrow25\rightarrow1$) as the set to set distance generator. The structure is summarized into Table~\ref{tab:s_strucutre}

\begin{table}[H]
    \begin{center}
    \scalebox{1}{
    \begin{tabular}{c|c| c }
                           
        \Xhline{2\arrayrulewidth}
    
            \bf{layer name}
          & \bf{output size}
          & \bf{operation parameter}
         
        \\
       \toprule\bottomrule
        linear1      & $25$       &$625\rightarrow25$                 
        \\
       \hline 
       1D batch norm & $25$       &$25$
        \\
       \hline 
       relu          &$25$        &-
        \\
       \hline 
       dropout       &$25$        & $p=0.5$
        \\
       \hline
       linear2      &$1$          &$25\rightarrow1$              
        \\
       \hline 
      1D batch norm &$1 \times 1$  &1

      \\
    \Xhline{2\arrayrulewidth}
    \end{tabular}
    }
    \end{center}
    \caption{The stucture of the Set to Set Distance Generator.}
    \vspace{0.5em}
    \label{tab:s_strucutre}
\end{table}

\subsection{Extra Visualizations and Discussion}
\begin{figure}[H]
\begin{center}
  \subfigure[]{\includegraphics[width=0.3\linewidth]{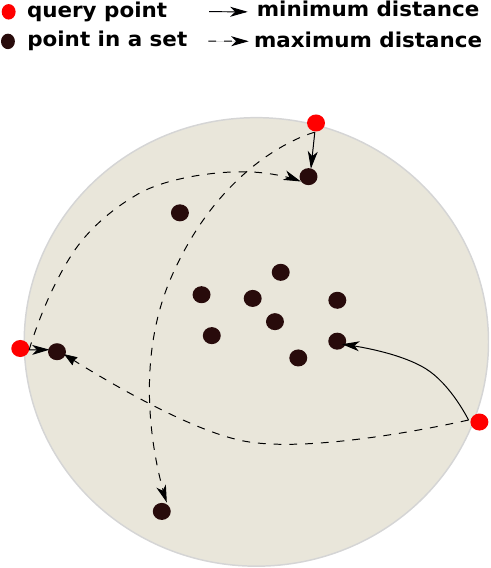}\label{fig:min-max}}%
  \subfigure[]{\includegraphics[width=0.3\linewidth]{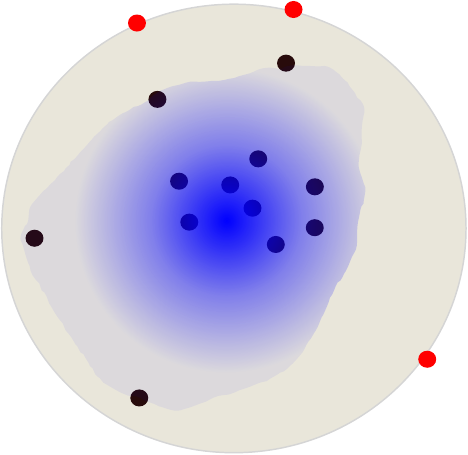}\label{fig:APP2S1}}%
	\subfigure[]{\includegraphics[width=0.3\linewidth]{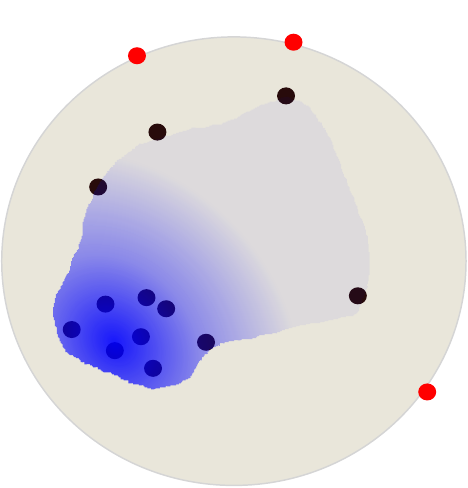}\label{fig:APP2S2}}%

\end{center}
  \caption{(a): The P2S distances based on the minimum and maximum distances are sensitive to outliers and ignore the distribution of the points in the set to a great degree.  (b) and (c): Our proposed point to set distance is bounded between the infimum and the supremum and, it is also non-linear due to the weighted-sum. It covers the  distribution of the individual sample in the set and adapts based on the relationship between the sample distribution and overall set distribution.}
\label{fig:p2s}
\end{figure}

We also provide extra visualizations to show that the APP2S will adapt depending on the constellation of the points in a set. Fig.~\ref{fig:p2s} shows that in both cases~\ref{fig:APP2S1} and~\ref{fig:APP2S2}, APP2S assigns larger weights (dark blue area) to the points that are closer to the center of the cluster, while smaller weights (light blue) to the outliers.

\noindent{\bf{Our P2S.}} The existing P2S distance metrics (\ie, the min and max distances discussed in Preliminary) only consider the lower bound and upper bound of P2S distance, thereby ignoring the distribution of the samples of the set to a great degree. Furthermore, such metrics are very sensitive to the outliers in the set (see Fig.~\ref{fig:min-max}). Our proposed adaptive P2S distance is a more flexible metric and able to adapt based on the distribution of the samples in the set. See Fig.~\ref{fig:APP2S1} and~\ref{fig:APP2S2} for an example, the measurement from our proposed metric is more flexible than the existing ones. Note that the weight (\ie, $w_{ij}$) generated by our method is distance-dependent. This is due to the way we model the problem using the tangent space of the hyperbolic space. To see this, recall that the norm of projected sample vector in support-class, which is the input of the relation generator, is indeed  the geodesic distance between the associated support vector and the query vector on the manifold (\ie, $\|{\tilde{\Vec{s}}}_{ij}^{{r}} \| = d_c(\bar{\Vec{q}}, {{\Vec{s}}}_{ij}^{{r}})$).

\noindent{\textbf{RelationNet.}} Our relation generator resembles the RelationNet. However, instead of computing the relation score between the prototype and the query, our relation generator computes the relation score between each support sample and its corresponding class-signature, further used as the adaptive factors for our point-to-set distance. 

\noindent{\textbf{DN4.}} The distance in DN4 resembles the point to set distance in our work. However and in contrast to DN4, our point to set distance is adaptive, while that in DN4 is fixed weighted summation.

\bibliography{aaai22.bib}

\begin{thebibliography}{52}
\providecommand{\natexlab}[1]{#1}

\bibitem[{Antoniou, Edwards, and Storkey(2019)}]{antoniou2019train}
Antoniou, A.; Edwards, H.; and Storkey, A. 2019.
\newblock How to train your MAML.
\newblock In \emph{International Conference on Learning Representations}.

\bibitem[{Bertinetto et~al.(2018)Bertinetto, Henriques, Torr, and
  Vedaldi}]{bertinetto2018meta}
Bertinetto, L.; Henriques, J.~F.; Torr, P.; and Vedaldi, A. 2018.
\newblock Meta-learning with differentiable closed-form solvers.
\newblock In \emph{International Conference on Learning Representations}.

\bibitem[{Carion et~al.(2020)Carion, Massa, Synnaeve, Usunier, Kirillov, and
  Zagoruyko}]{carion2020end}
Carion, N.; Massa, F.; Synnaeve, G.; Usunier, N.; Kirillov, A.; and Zagoruyko,
  S. 2020.
\newblock End-to-end object detection with transformers.
\newblock In \emph{European Conference on Computer Vision}, 213--229. Springer.

\bibitem[{Chen et~al.(2019)Chen, Liu, Kira, Wang, and Huang}]{chen2019closer}
Chen, W.-Y.; Liu, Y.-C.; Kira, Z.; Wang, Y.-C.~F.; and Huang, J.-B. 2019.
\newblock A Closer Look at Few-shot Classification.
\newblock In \emph{International Conference on Learning Representations}.

\bibitem[{Conci and Kubrusly(2018)}]{Conci2018DistanceBS}
Conci, A.; and Kubrusly, C.~S. 2018.
\newblock Distance Between Sets - A survey.

\bibitem[{Deng et~al.(2009)Deng, Dong, Socher, Li, Li, and
  Fei-Fei}]{deng2009imagenet}
Deng, J.; Dong, W.; Socher, R.; Li, L.-J.; Li, K.; and Fei-Fei, L. 2009.
\newblock Imagenet: A large-scale hierarchical image database.
\newblock In \emph{IEEE Conference on Computer Vision and Pattern Recognition}.

\bibitem[{Doersch, Gupta, and Zisserman(2020)}]{doersch2020crosstransformers}
Doersch, C.; Gupta, A.; and Zisserman, A. 2020.
\newblock Crosstransformers: spatially-aware few-shot transfer.
\newblock \emph{arXiv preprint arXiv:2007.11498}.

\bibitem[{Fang, Harandi, and Petersson(2021)}]{Fang_2021_ICCV}
Fang, P.; Harandi, M.; and Petersson, L. 2021.
\newblock Kernel Methods in Hyperbolic Spaces.
\newblock In \emph{Proceedings of the IEEE/CVF International Conference on
  Computer Vision}.

\bibitem[{Fang et~al.(2021)Fang, Ji, Petersson, and Harandi}]{Fang_2021_WACV}
Fang, P.; Ji, P.; Petersson, L.; and Harandi, M. 2021.
\newblock Set Augmented Triplet Loss for Video Person Re-Identification.
\newblock In \emph{Proceedings of the IEEE/CVF Winter Conference on
  Applications of Computer Vision WACV}.

\bibitem[{Finn, Abbeel, and Levine(2017)}]{finn2017model}
Finn, C.; Abbeel, P.; and Levine, S. 2017.
\newblock Model-agnostic meta-learning for fast adaptation of deep networks.
\newblock In \emph{Proceedings of the 34th International Conference on Machine
  Learning-Volume 70}, 1126--1135.

\bibitem[{Flennerhag et~al.(2019)Flennerhag, Rusu, Pascanu, Visin, Yin, and
  Hadsell}]{flennerhag2019meta}
Flennerhag, S.; Rusu, A.~A.; Pascanu, R.; Visin, F.; Yin, H.; and Hadsell, R.
  2019.
\newblock Meta-Learning with Warped Gradient Descent.
\newblock In \emph{International Conference on Learning Representations}.

\bibitem[{Franceschi et~al.(2018)Franceschi, Frasconi, Salzo, Grazzi, and
  Pontil}]{franceschi2018bilevel}
Franceschi, L.; Frasconi, P.; Salzo, S.; Grazzi, R.; and Pontil, M. 2018.
\newblock Bilevel programming for hyperparameter optimization and
  meta-learning.
\newblock In \emph{International Conference on Machine Learning}, 1568--1577.
  PMLR.

\bibitem[{Ganea, B{\'e}cigneul, and Hofmann(2018)}]{ganea2018hyperbolic}
Ganea, O.; B{\'e}cigneul, G.; and Hofmann, T. 2018.
\newblock Hyperbolic neural networks.
\newblock In \emph{Advances in Neural Information Processing Systems}.

\bibitem[{Ganea(2019)}]{ganea2019non}
Ganea, O.-E. 2019.
\newblock \emph{Non-Euclidean Neural Representation Learning of Words, Entities
  and Hierarchies}.
\newblock Ph.D. thesis, ETH Zurich.

\bibitem[{Gidaris and Komodakis(2018)}]{gidaris2018dynamic}
Gidaris, S.; and Komodakis, N. 2018.
\newblock Dynamic few-shot visual learning without forgetting.
\newblock In \emph{IEEE Conference on Computer Vision and Pattern Recognition}.

\bibitem[{Gidaris and Komodakis(2019)}]{gidaris2019generating}
Gidaris, S.; and Komodakis, N. 2019.
\newblock Generating classification weights with gnn denoising autoencoders for
  few-shot learning.
\newblock In \emph{IEEE Conference on Computer Vision and Pattern Recognition}.

\bibitem[{He et~al.(2016)He, Zhang, Ren, and Sun}]{he2016deep}
He, K.; Zhang, X.; Ren, S.; and Sun, J. 2016.
\newblock Deep residual learning for image recognition.
\newblock In \emph{IEEE conference on Computer Vision and Pattern Recognition}.

\bibitem[{Hong et~al.(2021)Hong, Fang, Li, Zhang, Simon, Harandi, and
  Petersson}]{Hong_2021_CVPR_RAP}
Hong, J.; Fang, P.; Li, W.; Zhang, T.; Simon, C.; Harandi, M.; and Petersson,
  L. 2021.
\newblock Reinforced Attention for Few-Shot Learning and Beyond.
\newblock In \emph{Proceedings of the IEEE/CVF Conference on Computer Vision
  and Pattern Recognition}.

\bibitem[{Huttenlocher, Klanderman, and
  Rucklidge(1993)}]{huttenlocher1993comparing}
Huttenlocher, D.~P.; Klanderman, G.~A.; and Rucklidge, W.~J. 1993.
\newblock Comparing images using the Hausdorff distance.
\newblock \emph{IEEE Transactions on Pattern Analysis and Machine
  Intelligence}.

\bibitem[{Ibanez et~al.(2008)Ibanez, Audette, Yeo, Golland, Tustison, and
  Gee}]{ibanez2008use}
Ibanez, L.; Audette, M.; Yeo, B.; Golland, P.; Tustison, N.; and Gee, J. 2008.
\newblock The Use of Robust Local Hausdorff Distances in Accuracy Assessment
  for Image Alignment of Brain MRI.
\newblock \emph{Insight Journal}.

\bibitem[{Khrulkov et~al.(2020)Khrulkov, Mirvakhabova, Ustinova, Oseledets, and
  Lempitsky}]{khrulkov2020hyperbolic}
Khrulkov, V.; Mirvakhabova, L.; Ustinova, E.; Oseledets, I.; and Lempitsky, V.
  2020.
\newblock Hyperbolic image embeddings.
\newblock In \emph{IEEE/CVF Conference on Computer Vision and Pattern
  Recognition}.

\bibitem[{Kingma and Ba(2014)}]{kingma2014adam}
Kingma, D.~P.; and Ba, J. 2014.
\newblock Adam: A method for stochastic optimization.
\newblock \emph{arXiv preprint arXiv:1412.6980}.

\bibitem[{Lee et~al.(2019{\natexlab{a}})Lee, Na, Lee, and
  Hwang}]{lee2019learning}
Lee, H.; Na, D.; Lee, H.~B.; and Hwang, S.~J. 2019{\natexlab{a}}.
\newblock Learning to Generalize to Unseen Tasks with Bilevel Optimization.
\newblock \emph{arXiv preprint arXiv:1908.01457}.

\bibitem[{Lee et~al.(2019{\natexlab{b}})Lee, Maji, Ravichandran, and
  Soatto}]{lee2019meta}
Lee, K.; Maji, S.; Ravichandran, A.; and Soatto, S. 2019{\natexlab{b}}.
\newblock Meta-learning with differentiable convex optimization.
\newblock In \emph{IEEE Conference on Computer Vision and Pattern Recognition}.

\bibitem[{Li et~al.(2019{\natexlab{a}})Li, Eigen, Dodge, Zeiler, and
  Wang}]{li2019finding}
Li, H.; Eigen, D.; Dodge, S.; Zeiler, M.; and Wang, X. 2019{\natexlab{a}}.
\newblock Finding task-relevant features for few-shot learning by category
  traversal.
\newblock In \emph{IEEE/CVF Conference on Computer Vision and Pattern
  Recognition}.

\bibitem[{Li et~al.(2020)Li, Zhang, Li, and Fu}]{li2020adversarial}
Li, K.; Zhang, Y.; Li, K.; and Fu, Y. 2020.
\newblock Adversarial Feature Hallucination Networks for Few-Shot Learning.
\newblock In \emph{IEEE/CVF Conference on Computer Vision and Pattern
  Recognition}.

\bibitem[{Li et~al.(2019{\natexlab{b}})Li, Wang, Xu, Huo, Gao, and
  Luo}]{li2019revisiting}
Li, W.; Wang, L.; Xu, J.; Huo, J.; Gao, Y.; and Luo, J. 2019{\natexlab{b}}.
\newblock Revisiting local descriptor based image-to-class measure for few-shot
  learning.
\newblock In \emph{Proceedings of the IEEE/CVF Conference on Computer Vision
  and Pattern Recognition}, 7260--7268.

\bibitem[{Lifchitz et~al.(2019)Lifchitz, Avrithis, Picard, and
  Bursuc}]{lifchitz2019dense}
Lifchitz, Y.; Avrithis, Y.; Picard, S.; and Bursuc, A. 2019.
\newblock Dense classification and implanting for few-shot learning.
\newblock In \emph{Proceedings of the IEEE/CVF Conference on Computer Vision
  and Pattern Recognition}, 9258--9267.

\bibitem[{Liu et~al.(2020)Liu, Cao, Lin, Li, Zhang, Long, and
  Hu}]{liu2020negative}
Liu, B.; Cao, Y.; Lin, Y.; Li, Q.; Zhang, Z.; Long, M.; and Hu, H. 2020.
\newblock Negative margin matters: Understanding margin in few-shot
  classification.
\newblock In \emph{European Conference on Computer Vision}, 438--455. Springer.

\bibitem[{Lu, Ye, and Zhan(2021)}]{lu2021tailoring}
Lu, S.; Ye, H.-J.; and Zhan, D.-C. 2021.
\newblock Tailoring Embedding Function to Heterogeneous Few-Shot Tasks by
  Global and Local Feature Adaptors.
\newblock In \emph{Proceedings of the AAAI Conference on Artificial
  Intelligence}, volume~35, 8776--8783.

\bibitem[{Nichol, Achiam, and Schulman(2018)}]{nichol2018first}
Nichol, A.; Achiam, J.; and Schulman, J. 2018.
\newblock On first-order meta-learning algorithms.
\newblock \emph{arXiv preprint arXiv:1803.02999}.

\bibitem[{Oreshkin, L{\'o}pez, and Lacoste(2018)}]{oreshkin2018tadam}
Oreshkin, B.; L{\'o}pez, P.~R.; and Lacoste, A. 2018.
\newblock Tadam: Task dependent adaptive metric for improved few-shot learning.
\newblock In \emph{Advances in Neural Information Processing Systems}.

\bibitem[{Qiao et~al.(2019)Qiao, Shi, Li, Wang, Huang, and
  Tian}]{qiao2019transductive}
Qiao, L.; Shi, Y.; Li, J.; Wang, Y.; Huang, T.; and Tian, Y. 2019.
\newblock Transductive episodic-wise adaptive metric for few-shot learning.
\newblock In \emph{IEEE International Conference on Computer Vision}.

\bibitem[{Ravi and Larochelle(2016)}]{ravi2016optimization}
Ravi, S.; and Larochelle, H. 2016.
\newblock Optimization as a model for few-shot learning.
\newblock In \emph{International Conference on Learning Representations}.

\bibitem[{Ravichandran, Bhotika, and Soatto(2019)}]{ravichandran2019few}
Ravichandran, A.; Bhotika, R.; and Soatto, S. 2019.
\newblock Few-shot learning with embedded class models and shot-free meta
  training.
\newblock In \emph{IEEE International Conference on Computer Vision}.

\bibitem[{Ren et~al.(2018)Ren, Triantafillou, Ravi, Snell, Swersky, Tenenbaum,
  Larochelle, and Zemel}]{ren2018meta}
Ren, M.; Triantafillou, E.; Ravi, S.; Snell, J.; Swersky, K.; Tenenbaum, J.~B.;
  Larochelle, H.; and Zemel, R.~S. 2018.
\newblock Meta-Learning for Semi-Supervised Few-Shot Classification.
\newblock In \emph{International Conference on Learning Representations}.

\bibitem[{Rusu et~al.(2018)Rusu, Rao, Sygnowski, Vinyals, Pascanu, Osindero,
  and Hadsell}]{rusu2018meta}
Rusu, A.~A.; Rao, D.; Sygnowski, J.; Vinyals, O.; Pascanu, R.; Osindero, S.;
  and Hadsell, R. 2018.
\newblock Meta-Learning with Latent Embedding Optimization.
\newblock In \emph{International Conference on Learning Representations}.

\bibitem[{Shen et~al.(2021)Shen, Liu, Qin, Savvides, and
  Cheng}]{shen2021partial}
Shen, Z.; Liu, Z.; Qin, J.; Savvides, M.; and Cheng, K.-T. 2021.
\newblock Partial Is Better Than All: Revisiting Fine-tuning Strategy for
  Few-shot Learning.
\newblock In \emph{Proceedings of the AAAI Conference on Artificial
  Intelligence}, volume~35, 9594--9602.

\bibitem[{Simon et~al.(2020)Simon, Koniusz, Nock, and
  Harandi}]{simon2020adaptive}
Simon, C.; Koniusz, P.; Nock, R.; and Harandi, M. 2020.
\newblock Adaptive Subspaces for Few-Shot Learning.
\newblock In \emph{IEEE/CVF Conference on Computer Vision and Pattern
  Recognition}.

\bibitem[{Snell, Swersky, and Zemel(2017)}]{snell2017prototypical}
Snell, J.; Swersky, K.; and Zemel, R. 2017.
\newblock Prototypical networks for few-shot learning.
\newblock In \emph{Advances in Neural Information Processing Systems}.

\bibitem[{Su, Maji, and Hariharan(2020)}]{su2020does}
Su, J.-C.; Maji, S.; and Hariharan, B. 2020.
\newblock When does self-supervision improve few-shot learning?
\newblock In \emph{European Conference on Computer Vision}, 645--666. Springer.

\bibitem[{Sun et~al.(2019)Sun, Sun, Zhou, and Lv}]{sun2019hierarchical}
Sun, S.; Sun, Q.; Zhou, K.; and Lv, T. 2019.
\newblock Hierarchical attention prototypical networks for few-shot text
  classification.
\newblock In \emph{Proceedings of the 2019 Conference on Empirical Methods in
  Natural Language Processing and the 9th International Joint Conference on
  Natural Language Processing}, 476--485.

\bibitem[{Sung et~al.(2018)Sung, Yang, Zhang, Xiang, Torr, and
  Hospedales}]{sung2018learning}
Sung, F.; Yang, Y.; Zhang, L.; Xiang, T.; Torr, P.~H.; and Hospedales, T.~M.
  2018.
\newblock Learning to compare: Relation network for few-shot learning.
\newblock In \emph{IEEE Conference on Computer Vision and Pattern Recognition}.

\bibitem[{Tang et~al.(2020)Tang, Li, Peng, and Tang}]{tang2020blockmix}
Tang, H.; Li, Z.; Peng, Z.; and Tang, J. 2020.
\newblock BlockMix: meta regularization and self-calibrated inference for
  metric-based meta-learning.
\newblock In \emph{Proceedings of the 28th ACM International Conference on
  Multimedia}, 610--618.

\bibitem[{Vinyals et~al.(2016)Vinyals, Blundell, Lillicrap, Wierstra
  et~al.}]{vinyals2016matching}
Vinyals, O.; Blundell, C.; Lillicrap, T.; Wierstra, D.; et~al. 2016.
\newblock Matching networks for one shot learning.
\newblock In \emph{Advances in Neural Information Processing Systems}.

\bibitem[{Wah et~al.(2011)Wah, Branson, Welinder, Perona, and
  Belongie}]{wah2011caltech}
Wah, C.; Branson, S.; Welinder, P.; Perona, P.; and Belongie, S. 2011.
\newblock The caltech-ucsd birds-200-2011 dataset.

\bibitem[{Wang et~al.(2019)Wang, Chao, Weinberger, and van~der
  Maaten}]{wang2019simpleshot}
Wang, Y.; Chao, W.-L.; Weinberger, K.~Q.; and van~der Maaten, L. 2019.
\newblock Simpleshot: Revisiting nearest-neighbor classification for few-shot
  learning.
\newblock \emph{arXiv preprint arXiv:1911.04623}.

\bibitem[{Wang et~al.(2020)Wang, Xu, Liu, Zhang, and Fu}]{wang2020instance}
Wang, Y.; Xu, C.; Liu, C.; Zhang, L.; and Fu, Y. 2020.
\newblock Instance credibility inference for few-shot learning.
\newblock In \emph{Proceedings of the IEEE/CVF Conference on Computer Vision
  and Pattern Recognition}, 12836--12845.

\bibitem[{Wertheimer, Tang, and Hariharan(2021)}]{wertheimer2021few}
Wertheimer, D.; Tang, L.; and Hariharan, B. 2021.
\newblock Few-Shot Classification With Feature Map Reconstruction Networks.
\newblock In \emph{Proceedings of the IEEE/CVF Conference on Computer Vision
  and Pattern Recognition}, 8012--8021.

\bibitem[{Xu et~al.(2021)Xu, Fu, Liu, Wang, Li, Huang, Zhang, and
  Xue}]{xu2021learning}
Xu, C.; Fu, Y.; Liu, C.; Wang, C.; Li, J.; Huang, F.; Zhang, L.; and Xue, X.
  2021.
\newblock Learning Dynamic Alignment via Meta-filter for Few-shot Learning.
\newblock In \emph{Proceedings of the IEEE/CVF Conference on Computer Vision
  and Pattern Recognition}, 5182--5191.

\bibitem[{Ye et~al.(2020)Ye, Hu, Zhan, and Sha}]{ye2020few}
Ye, H.-J.; Hu, H.; Zhan, D.-C.; and Sha, F. 2020.
\newblock Few-shot learning via embedding adaptation with set-to-set functions.
\newblock In \emph{IEEE/CVF Conference on Computer Vision and Pattern
  Recognition}.

\bibitem[{Zhang et~al.(2020)Zhang, Cai, Lin, and Shen}]{zhang2020deepemd}
Zhang, C.; Cai, Y.; Lin, G.; and Shen, C. 2020.
\newblock DeepEMD: Few-Shot Image Classification with Differentiable Earth
  Mover's Distance and Structured Classifiers.
\newblock In \emph{IEEE/CVF Conference on Computer Vision and Pattern
  Recognition}.

\end{thebibliography}

\end{document}